\documentclass{article}

\usepackage[final]{neurips_2025}

\usepackage[utf8]{inputenc} 
\usepackage[T1]{fontenc}    
\usepackage{hyperref}       
\usepackage{url}            
\usepackage{booktabs}       
\usepackage{amsfonts}       
\usepackage{amsmath}
\usepackage{nicefrac}       
\usepackage{microtype}      
\usepackage{xcolor}         

\usepackage{graphicx}
\usepackage{subfigure}
\usepackage{algorithmic}
\usepackage{algorithm}

\title{Improving Regret Approximation for Unsupervised Dynamic Environment Generation}

\author{%
  Harry Mead \\
  University of Oxford \\
  \And
  Bruno Lacerda \\
  University of Oxford \\
  \And 
  Jakob Foerster \\
  University of Oxford \\
  \And
  Nick Hawes \\
  University of Oxford \\
}

\begin{document}

\maketitle

\begin{abstract}
  Unsupervised Environment Design (UED) seeks to automatically generate training curricula for reinforcement learning (RL) agents, with the goal of improving generalisation and zero-shot performance. However, designing effective curricula remains a difficult problem, particularly in settings where small subsets of environment parameterisations result in significant increases in the complexity of the required policy. Current methods struggle with a difficult credit assignment problem and rely on regret approximations that fail to identify challenging levels, both of which are compounded as the size of the environment grows. We propose Dynamic Environment Generation for UED (DEGen) to enable a denser level generator reward signal, reducing the difficulty of credit assignment and allowing for UED to scale to larger environment sizes. We also introduce a new regret approximation, Maximised Negative Advantage (MNA), as a significantly improved metric to optimise for, that better identifies more challenging levels. We show empirically that MNA outperforms current regret approximations and when combined with DEGen, consistently outperforms existing methods, especially as the size of the environment grows. We have made all our code available here: \url{https://github.com/HarryMJMead/Dynamic-Environment-Generation-for-UED}.
\end{abstract}

\section{Introduction}

Deep Reinforcement Learning (RL) has been effective in training highly-capable agents in a number of different challenging settings, such as in real-world robotics applications \cite{akkaya2019solving, andrychowicz2020learning, rudin2022learning, kaufmann2023champion}, or games such as Go \cite{silver2016mastering}, Chess \cite{silver2017mastering}, Starcraft \cite{vinyals2019grandmaster} and Dota \cite{berner2019dota}. However, these deep-RL agents tend to exhibit poor generalisation when transferred to tasks or environments with only small changes to those used to train on \cite{zhang2018dissection, cobbe2019quantifying}. 

In order to address this lack of robustness, domain-randomisation (DR), training over a diversity of environment parameterisations, has proven successful in a number of applications. However, DR relies on random parameterisations resulting in useful training examples, and in complex environments this may not be the case. Automated Curriculum Learning (ACL) \cite{florensa2018automatic, portelas2020teacher} methods aim to produce adaptive curricula for training that ensure the generation of useful training examples whilst maintaining a sufficiently diverse distribution over these environment parameterisations. These methods have shown success over naive domain-randomisation approaches \cite{portelas2020automatic, narvekar2020curriculum}.

However, manually designing a suitable curriculum for learning may in itself be a challenge, whilst also limiting the capacity for open-ended learning \cite{wang2019paired, wang2020enhanced}. Recent work has focused on Unsupervised Environment Design (UED) \cite{dennis2020emergent}, which has emerged as a widely applicable curriculum design method as no prior environment knowledge is required. In the UED literature, each parameterisation of the environment is referred to as a level, and so UED frames the curriculum design problem as the interaction between a teacher agent designing levels and a student agent training on these levels. The majority of existing work focuses on maximising student regret \cite{dennis2020emergent, jiang2021prioritized, parker2022evolving, chung2024adversarial}, as prior work \cite{dennis2020emergent} has shown that if the student and teacher reach a Nash equilibrium of a minimax regret game, the student must necessarily be able to solve all solvable environments. However, computing regret is intractable for many complex tasks, so these methods require \textit{regret approximations}. 

\begin{figure*}[t]
  \centering
  \subfigure[No Key Required]{
    \includegraphics[width=0.32\textwidth]{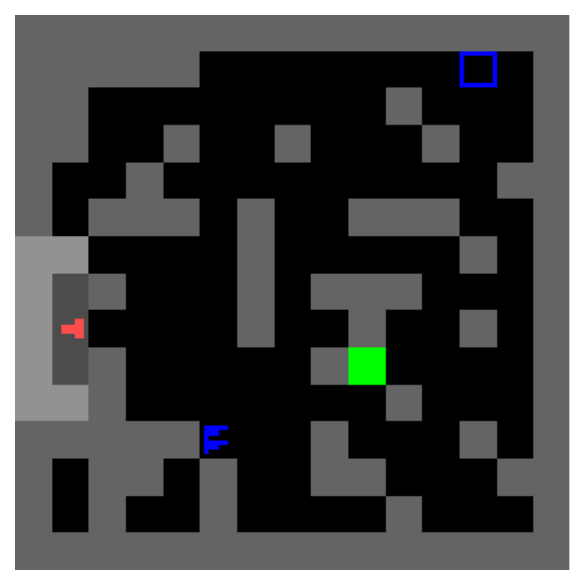}
    \label{fig:Random_Key_Level_No_Key}
  }
  \subfigure[Key Required]{
    \includegraphics[width=0.32\textwidth]{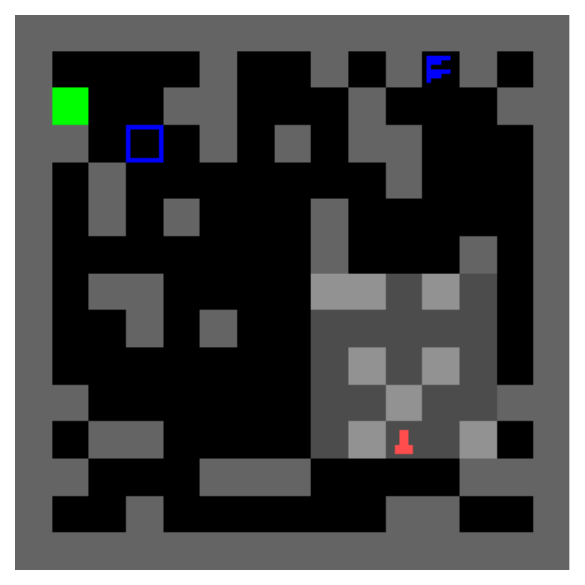}
    \label{fig:Random_Key_Level_Key}
  }
  \caption{Examples of two possible randomly generated levels. In the first, the agent (red triangle) can simply navigate to the goal (green square), whereas in the second, it is required to first obtain the key (two blue triangles) in order to unlock the door (blue unfilled square) blocking the path to the goal}
  \label{fig:Random_Generation_Comparison}
\end{figure*}

Generally UED methods can be categorised as either relying on a learnt level generator \cite{dennis2020emergent, mediratta2023stabilizing, azad2023clutr}, or a curation process that selects and replays levels from a randomly generated set \cite{jiang2021replay, jiang2021prioritized, parker2022evolving}. Existing UED methods have focused on environments such as minigrid \cite{dennis2020emergent, chung2024adversarial} or bipedal walker \cite{parker2022evolving}, where there is a relatively smooth transition in difficulty between levels. However, there are many environments where a small subset of paramaterisations may induce a step-change in difficulty for the student. For example, in the level shown in Figure \ref{fig:Random_Key_Level_No_Key}, the addition of the door and key to the level generation have no effect on the difficulty of the level for the student agent. The key can simply be ignored and the door acts as any other wall, and so the agent is able to navigate to the goal directly. However, in Figure \ref{fig:Random_Key_Level_Key}, the door blocks the agent's path to the goal, so it is necessary for the agent to first find the key before being able to unlock the door and reach the goal. In this example, these key-requiring mazes represent a very small subset of possible random levels, but they also represent the levels potentially most difficult to learn. Thus, for UED to be effective in environments such as these, it is necessary for methods to identify and train on this more challenging subset of levels.

Whilst replay-based methods are sufficient in small environments, the challenge of sampling and identifying more difficult subsets of levels is amplified as the size of the environment increases. Due to their reliance on random level generation, we show that replay methods fail in larger environments, and thus it is necessary to use learnt level generators. However, training a generator that generates a full level prior to student rollouts presents a challenging credit assignment task, given the long time horizon and sparse rewards. In order to address these challenges with learnt level generation, we propose Dynamic Environment Generation for Unsupervised Environment Design (DEGen). Our method involves dynamically generating the environment as the student agent explores the level, enabling a much denser teacher reward signal, and reducing the difficulty of credit assignment. 

However, we show in this work that current regret approximations are insufficient, both for identifying these most difficult subsets of levels and for use in training level generators. We propose Maximised Negative Advantage (MNA) as a more effective regret approximation and show substantial empirical performance improvements over existing regret approximation metrics. Using MNA, we show that DEGen performs substantially better than existing generators that rely on full level generation upfront. We show that DEGen is capable of matching or exceeding the performance of existing replay methods in small environments, unlike previous learnt generators, but performs substantially better as the size of the environment increases. We also show that MNA consistently improves performance over current regret approximations for all UED methods.

\newpage

Our contributions are:
\begin{itemize}
    \item We introduce \textit{Dyanamic Environment Generation for UED} (DEGen) as a new method of environment generation, showing performance improvements over existing learnt generators in small environments and significant performance improvements over all methods in larger environments.
    \item We introduce \textit{Maximised Negative Advantage} (MNA) as a new regret approximation and show substantial improvements over existing metrics.
\end{itemize}

\section{Background}

\subsection{Unsupervised Environment Design}
Given a specific environment, we can model a level as a Partially Observable Markov Decision Process (POMDP). POMDPs can be defined by a tuple $\langle S, A, O, \mathcal{T}, \mathcal{I}, \mathcal{R}, \rho_0, \gamma \rangle$, where $S$, $A$ and $O$ are the set of states, actions and observations respectively, $\mathcal{T}: S \times A \rightarrow S$ is the transition function, mapping a state-action pair $(s_t, a_t)$ to the subsequent state $s_{t+1}$, $\mathcal{I}: S \rightarrow O$ is the observation function that maps a given state to an observation, $\mathcal{R}: S \times A \rightarrow \mathbb{R}$ is the reward function, $\rho_0$ is the distribution over initial states, and $\gamma$ is the discount factor.

In order to extend this formulation to the framework of UED, the Underspecified POMDP (UPOMDP) is introduced \cite{dennis2020emergent}, defined by the tuple $\mathcal{M} = \langle S, A, O, \mathcal{T^M}, \mathcal{I^M}, \mathcal{R^M}, \rho_0^\mathcal{M}, \gamma \rangle$. The UPOMDP formulation introduces $\Theta$, the set of all possible free environment parameters $\theta$, for which a specific $\theta$ results in the environment configuration defined by the POMDP $\mathcal{M}_\theta$ with the transition, state and reward functions $\mathcal{T}^\theta, \mathcal{I}^\theta, \mathcal{R}^\theta$ and the initial state distribution $\rho_0^\theta$. 

Generally, the UED objective is to identify training levels that maximise the student's regret, given the current student policy $\pi$. The regret is defined as: 
\begin{equation} \label{eqn:Optimal_Regret}
    \text{Regret}(\pi, \theta) = - U(\pi, \theta) + U(\pi^*_\theta, \theta)
\end{equation}
where $\pi^*_\theta$ is the optimal policy given $\theta$, and $U(\pi, \theta) = \mathbb{E}_{\pi, \mathcal{M}_\theta} \left[ \sum_{t=0}^T \gamma^t r_t\right]$, or the expected discounted return of the policy $\pi$. As such, UED can be framed as a two player minimax regret game:
\begin{equation} \label{eqn:Minimax_Regret}
    \min_{\pi \in \Pi} \max_{\theta \in \Theta} \text{Regret}(\pi, \theta).
\end{equation}
By framing UED as this regret-based minimax game, if the environment satisfies the reward conditions outlined in \cite{dennis2020emergent}, we can guarantee that if the student and teacher policies reach a Nash equilibrium, then the student policy must necessarily be capable of solving all solvable levels.

\subsection{Existing UED Methods}
Whilst the optimal objective shown in Equation \ref{eqn:Optimal_Regret} has robustness guarantees, in practise, it is infeasible for UED given $\pi^*_\theta$ is required. UED methods such as PAIRED \cite{dennis2020emergent} or CLUTR \cite{azad2023clutr} introduce an additional antagonist agent, and regret is approximated as the difference between the performance of the antagonist and student policies. Both these methods rely on RL-trained teacher, where the teacher aims to maximise the performance difference between the antagonist and the student. However, these RL-trained teachers tend to struggle with maintaining diversity over training environments \cite{jiang2021replay}. Some techniques have shown to improve performance, such as behaviour cloning between the antagonist and protagonist and the use of high entropy coefficients \cite{mediratta2023stabilizing}. However these RL-based methods still tend to be outperformed by replay-based methods.

Rather than relying on a learnt generator, PLR \cite{jiang2021prioritized}, relies on maintaining a replay buffer of high-regret levels that have been sampled from a random generator. PLR alternates between sampling new random levels and replaying previously sampled levels. PLR relies on a score function to approximate regret, with the two most commonly used score functions being \textit{Positive Value Loss} (PVL) and \textit{Maximum Monte Carlo} (MaxMC) \cite{jiang2021replay, parker2022evolving}.

PVL is defined as 
\begin{equation}
\label{eqn:PVL}
    \dfrac{1}{T} \sum_{t=0}^T\left( \max(0, \sum_{k=t}^T (\lambda \gamma)^{k-t} \delta_k) \right)
\end{equation}
where $\gamma$ is the discount factor, $\lambda$ is from the Generalised Advantage Estimator \cite{schulman2015high} and $\delta_t$ is the 1-step TD-error at timestep $t$. We can view PVL as approximating regret as the average advantage, but with the advantage clipped at 0. As such, maximising PVL can be seen as effectively maximising states where the student does better than expected, which intuitively appears to be mismatched with the regret objective. Despite this, empirically, PVL has been shown to be effective. 

MaxMC approximates regret using the maximum achieved return ($R_{max}$) on a given level, and is defined as 
\begin{equation}
\label{eqn:MaxMC}
    \dfrac{1}{T} \sum_{t=0}^T \left( R_{max} - \hat{V}(s_t) \right)
\end{equation}
where $\hat{V}(s_t)$ is the learnt value function approximation for the value of the current policy at state $s_t$. MaxMC appears a more intuitive approximation for regret than PVL. However by relying on a Monte Carlo approximation for regret, MaxMC requires a sufficient number of trials in a level such that $R_{max}$ is a good approximation for the optimal return. Additionally, MaxMC can only be used in environments where reward is obtained at the final step, as $R_{max}$ is dependent on the full episode reward.

Whilst PLR has been shown to be effective in a number of domains, relying on random level generation necessarily means that no insight is gained from past levels to influence future levels generation. ACCEL \cite{parker2022evolving} addresses this by augmenting PLR such that new levels are generated by mutating existing levels previously in the replay buffer. This evolutionary approach enables some capacity for learning from previously identified high-regret levels, but still relies on random mutations for new level generation.

An alternate approach for UED is Sampling for Learnability (SFL) \cite{rutherford2024no}. Similarly to PLR, SFL relies on sampling a set of randomly generated levels and selecting those with the highest scores for training. However, rather than this score metric approximating regret, SFL aims to train on levels with high learnability, defined as $p (1 -p)$, where $p$ is the success rate of the current policy on the sampled level.

\section{Dynamic Environment Generation} \label{sec:DEGen}

\begin{algorithm}[tb]
   \caption{DEGen}
   \label{alg:DEGen}
\begin{algorithmic}
   \STATE {\bfseries Initialise:} student policy $\pi_{\phi_1}$, generator policy $\Lambda_{\phi_2}$
   \WHILE{not converged}
       \STATE \textcolor{blue}{// Sample $N$ trajectories}
       \FOR{$n \in 1:N$}
           \STATE Initialise empty level
           \STATE \textcolor{blue}{// Take $T$ student steps}
           \FOR{$t_s \in 1:T$}
               \STATE \textcolor{blue}{// Generate partial level}
               \STATE Sample $\Lambda$ actions to generate section of level that has been observed but not generated
               \STATE \textcolor{blue}{// Take student action}
               \STATE Sample $\pi$ action
           \ENDFOR
           \STATE compute score using student trajectory $\tau_s$
           \STATE assign reward to generator trajectory $\tau_g$
       \ENDFOR
       \STATE Update $\phi_1$ according to sampled student trajectories
       \STATE Update $\phi_2$ according to sampled generator trajectories
   \ENDWHILE
\end{algorithmic}
\end{algorithm}

Existing work \cite{jiang2021replay, parker2022evolving, mediratta2023stabilizing} has shown that replay-based UED methods generally outperform methods relying on a learnt generator. This can be attributed to a number of factors, but primarily, level generation presents difficulties for learning. The level generation learning problem has a long time horizon and sparse rewards, and so credit assignment is challenging. Current UED have focused on relatively small environments where it is feasible to sample useful training levels from random level generation. However, as the size of the environment grows, and especially with environments with features that can add complexity, it becomes more difficult to sample useful levels. Therefore, it becomes necessary to use a learnt generator instead.


\begin{figure}[ht]
\vspace{0.2cm}
\begin{center}
\centering
\includegraphics[width=0.2\textwidth]{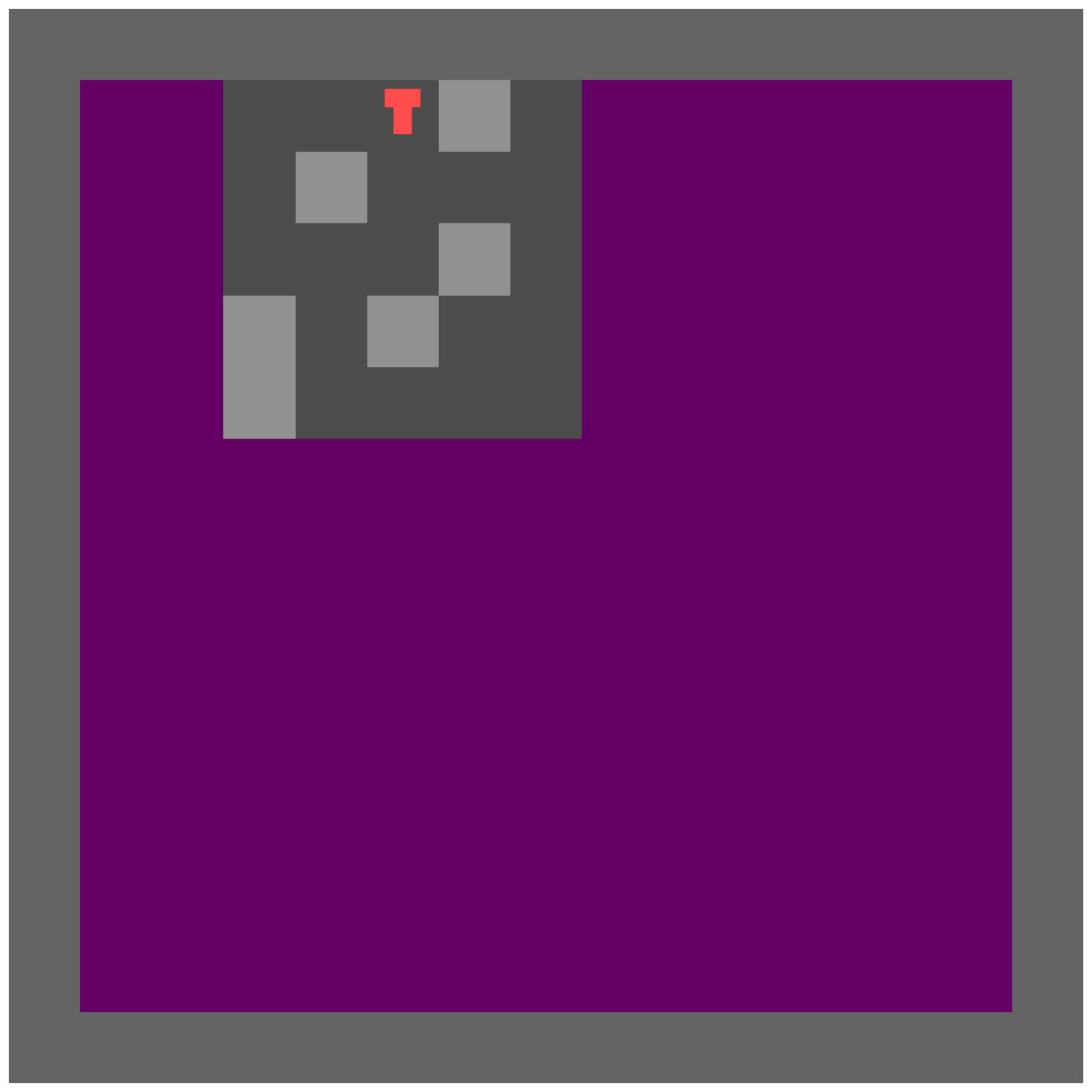}
\includegraphics[width=0.2\textwidth]{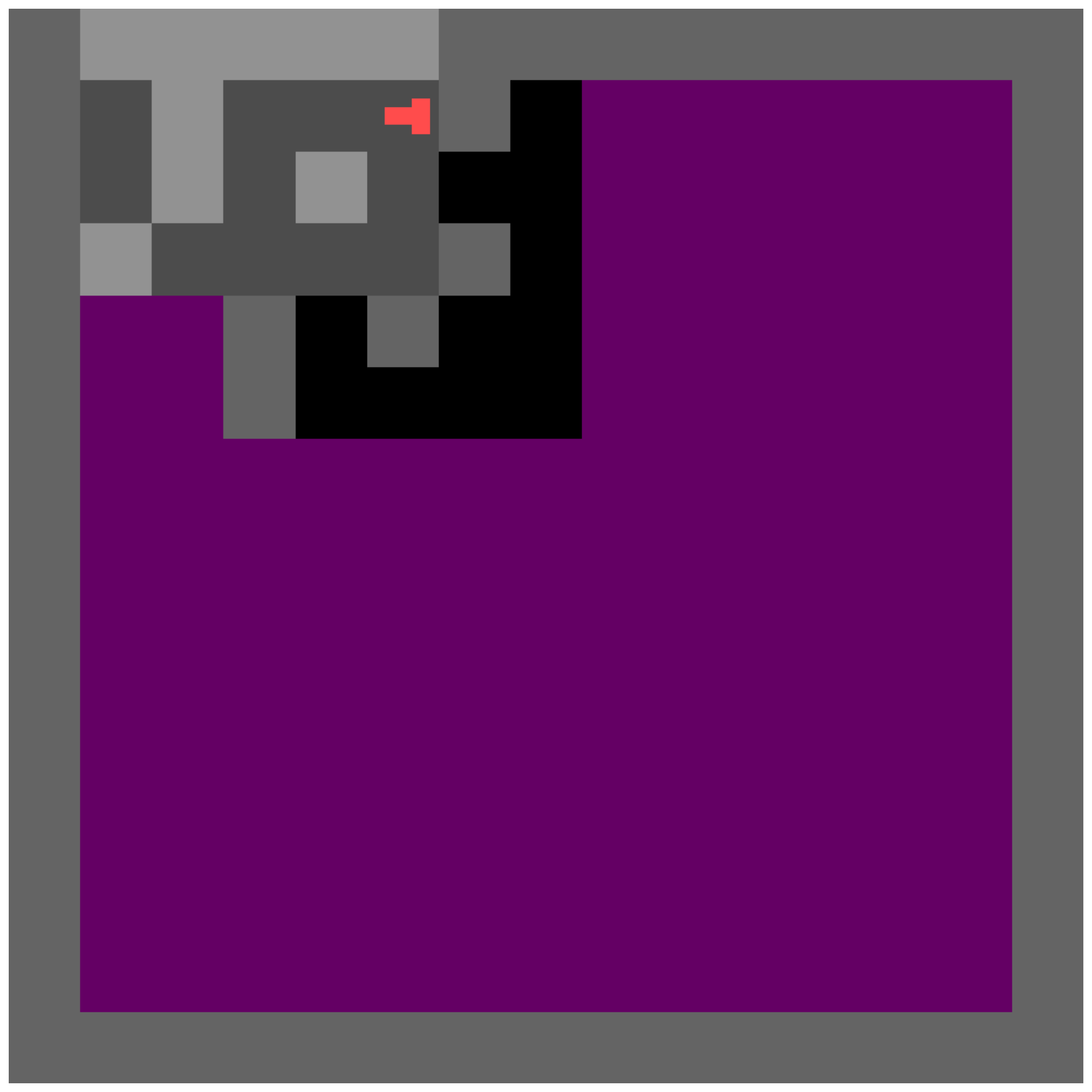}
\includegraphics[width=0.07\textwidth]{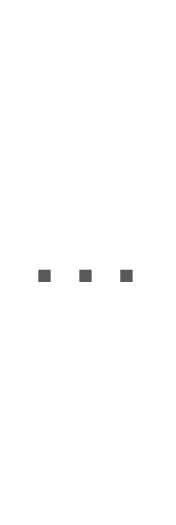}
\includegraphics[width=0.2\textwidth]{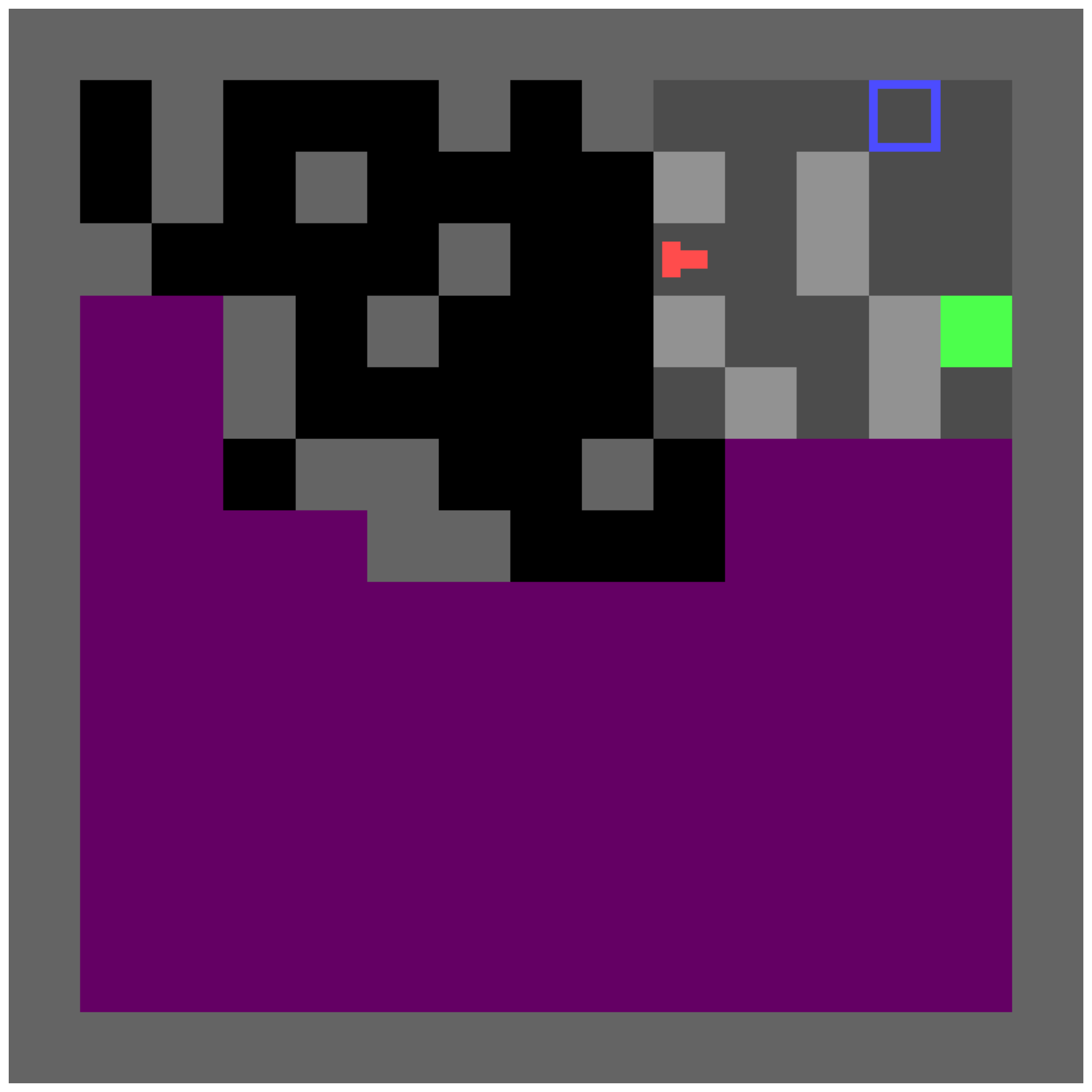}
\includegraphics[width=0.07\textwidth]{graphics/SPACING.pdf}
\includegraphics[width=0.2\textwidth]{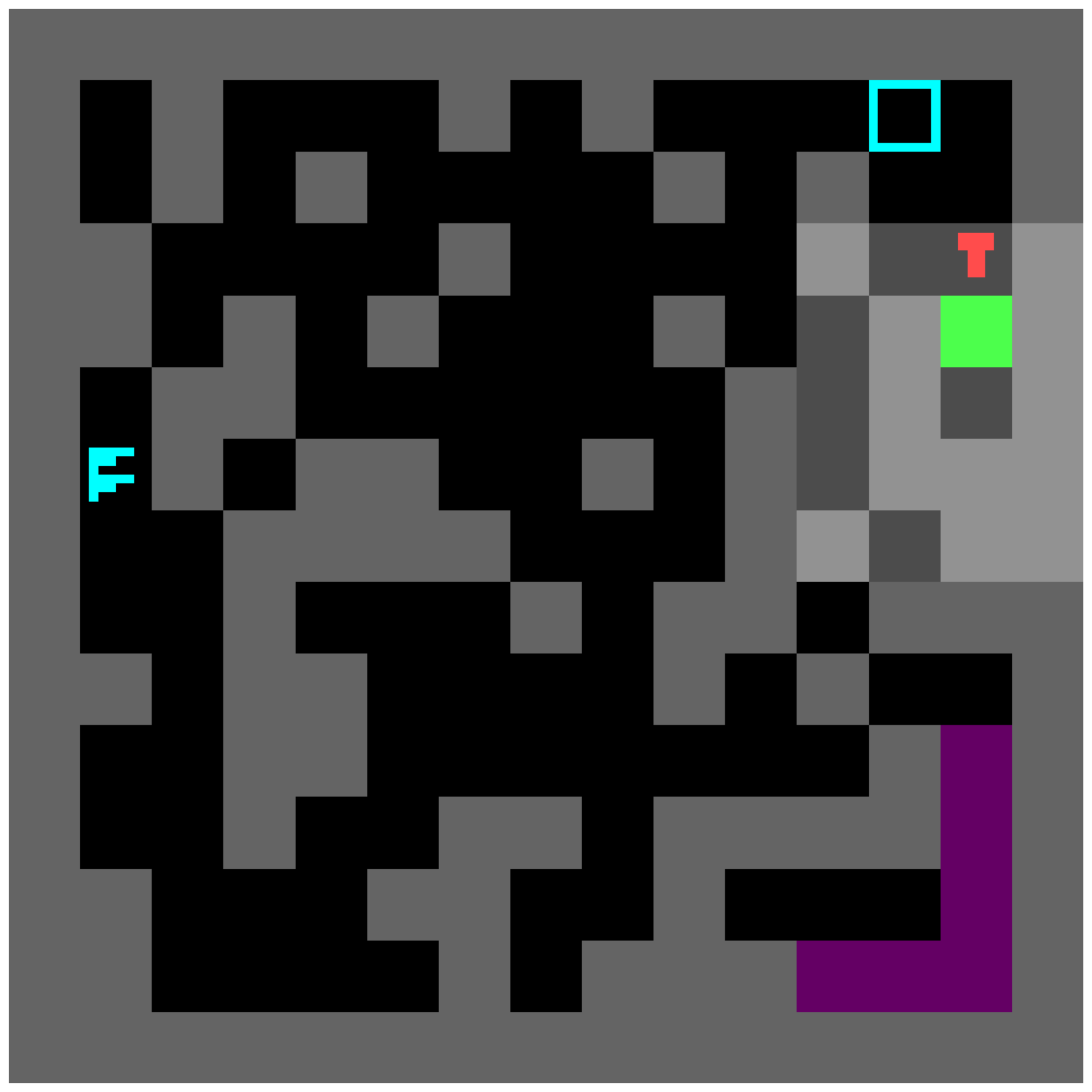}
\caption{Illustration of how DEGen generates the level as the student agent explores the level, with purple colouring indicating areas that are yet to be observed.}
\label{fig:DEGen_Image}
\end{center}
\end{figure}

In order to address this, whilst also reducing the challenges inherent to RL-based learnt generators, we propose Dynamic Environment Generation (DEGen), outlined in Algorithm \ref{alg:DEGen}. Rather than simply generating the entire environment level initially, we exploit the partial observability of the student agent and instead only generate those parts of the level the student observes, as illustrated in Figure \ref{fig:DEGen_Image}. This allows for a much denser reward signal, reducing the difficulty of credit assignment.

If we consider the PVL and MaxMC regret approximations shown in Equations \ref{eqn:PVL} and \ref{eqn:MaxMC} respectively, both can be written in the form
\begin{equation}
    \text{Regret} \approx \dfrac{1}{T}\sum_{t=0}^T G_t,
\end{equation}
where we approximate regret as the mean of some value $G_t$ across the student trajectory. If the entire level is generated initially, this approximation must simply be assigned as the reward for the last step in the level generation trajectory, resulting in a very sparse reward signal for the generator. However, if the level is generated as the student trajectory unrolls, we are able to assign a much denser generator reward. If we have the function $t_s = \mathcal{T}(t_g)$ that maps the generator timestep $t_g$ to the student timestep $t$, we can instead assign the generator reward ($r_g$) at timestep $t_g$ as
\begin{equation}
\label{eqn:Dense_Teacher_Rewards}
    r_{t_g} = \dfrac{1}{T}\sum_{t=\mathcal{T}(t_g)}^{\mathcal{T}(t_g+1)-1} G_t.
\end{equation}


This increased reward density reduces some of the difficulty of the credit assignment challenge for training the level generator. Additionally, DEGen reduces noise in the credit assignment by only generating parts of the level the student observers. If the entire level is generated initially, it is likely that some parts of the level will never be observed by the student, and so have no effect on the score of the level. As such, these actions only serve to add additional noise to the credit assignment task, an issue which is negated by DEGen. 

Another issue present in existing RL-based generators is a lack of diversity in generated levels. Previous work \cite{mediratta2023stabilizing} has shown that a high entropy coefficient can reduce this issue, however we still observed reduced level diversity with existing RL-based generators. As the student policy $\pi$ is stochastic, by generating the level based on where the student explores, we introduce an greater degree of stochasticity in the level generation, which increases the diversity of generated levels. We also found that introducing some additional randomness in level generation - specifically randomly initialising the starting location of the student agent - improved training level diversity, and zero shot agent performance. 

\section{Maximised Negative Advantage} \label{sec:MNA}

Whilst the PVL and MaxMC regret approximations have shown success in a number of domain, there are flaws with both metrics in relation to generating challenging levels. For PVL, high positive advantage will result in a high PVL score, whereas generally, more challenging levels would tend to be more difficult than expected, and so more likely to result in high negative advantage. For MaxMC, high scores require at least one high return rollout, and again, for the most challenging levels, this is unlikely to occur. Whilst PVL and MaxMC have shown success with use in replay-based methods, we observed poor performance when these metrics were directly optimised for, shown in Appendix \ref{app:DEGen_Regret_Approx_Compare}. We instead propose Maximised Negative Advantage (MNA) as a more suitable metric.

Consider the policy $\pi$ with the value-function $V(s)$. We are aiming for an approximation of Equation \ref{eqn:Optimal_Regret}, requiring an approximation for both $U(\pi_\theta^*, \theta)$ and $U(\pi, \theta)$. If we assume a deterministic environment, given a trajectory of length $T$, if $V(s)$ is the true value function, we can lower bound $U(\pi_\theta^*, \theta)$

\begin{equation} \label{eqn:Regret_Lower_Bound}
    U(\pi_\theta^*, \theta) \geq 
\max\left(
\begin{aligned}
&V(s_0), \\
&\gamma V(s_1) + r(s_0, a_0), \\
&\vdots \\
&\gamma^T V(s_T) + \sum_{k=0}^{T-1} \gamma^k r(s_k, a_k)
\end{aligned}
\right).
\end{equation}

We label this maximum over value functions 
\begin{equation}
    V^{\max} _n(s_t) = 
\max\left(
\begin{aligned}
&V(s_t), \\
&\vdots \\
&\gamma^n V(s_{t+n}) + \sum_{k=t}^{t+n-1} \gamma^{k-t} r(s_k, a_k)
\end{aligned}
\right).
\end{equation}

Given this, as the true value function $V(s_0)$ gives us the expected performance of the current policy, and the maximum over value functions $V^{\max} _T(s_0)$ lower bounds the performance of the optimal policy, we can lower bound the regret as
\begin{equation}
    \text{Regret} \geq -V(s_0) + V^{\max} _T(s_0).
\end{equation}

However, in practise, the exact value function $V(s)$ will generally be unknown, and instead must be approximated with a learnt value function $\hat{V}(s)$. As this learnt value function may overestimate the value of the state, the inequality in Equation \ref{eqn:Regret_Lower_Bound} does not necessarily hold when $V(s)$ is replaced with $\hat{V}(s)$. Additionally, the $\hat{V}(s_0)$ approximation for $U(\pi, \theta)$ will be biased, being a learnt value approximation. We can reduce the likelihood of $\hat{V}^{\max} _T(s_0)$ exceeding $U(\pi_\theta^*, \theta)$ by instead using the approximation $\hat{V}^{\max} _n(s_0)$, where $n<T$, although this instead results in a potentially overly conservative approximation. Similarly, we can reduce the bias of our approximation for $U(\pi, \theta)$ by instead using the approximation $\gamma^n V(s_{n}) + \sum_{k=t}^{n-1} \gamma^{k} r(s_k, a_k)$, however this then introduces greater variance. We therefore define the n-step regret approximation at timestep $t$ as 
\begin{equation} \label{eqn:n_step_regret}
    \hat{G}_t^{(n)} = -\Bigl( \gamma^n V(s_{t+n}) + \sum_{k=t}^{t+n-1} \gamma^{k-t} r(s_k, a_k) \Bigr) + \hat{V}^{\max} _n(s_t)
\end{equation}
and we note the similarity between this regret approximation and the negative n-step advantage estimation \cite{schulman2015high}. In order to balance both the bias and variance of the $U(\pi, \theta)$ approximation, and the conservativeness of the $U(\pi_\theta^*, \theta)$ approximation, in line with the Generalised Advantage Estimator \cite{schulman2015high}, we introduce the regret approximation
\begin{equation}
    \hat{G}_t^{\lambda} = (1-\lambda)\sum_{n=0}^\infty \lambda^n \hat{G}_t^{(n)}.
\end{equation}

Empirically, we find that rather than just approximating regret at the first state, using the mean regret approximation was a more effective metric
\begin{equation} \label{eqn:MNA_No_Solve}
    \dfrac{1}{T} \sum_{t=0}^T\hat{G}_t^{\lambda}.
\end{equation}
We show empirically that this regret approximation is much suitable optimisation metric for learnt generators, whilst also showing improved performance when used in replay-based methods.

\subsection{Solvability}

Whilst the metric in Equation \ref{eqn:MNA_No_Solve} does allow for more challenging levels to be sampled than existing metrics, the reliance on the learnt value function $\hat{V}(s)$ to determine maximum possible performance does present issues with ensuring level solvability. If the environment satisfies the reward conditions \cite{dennis2020emergent} that ensure the teacher-student regret Nash equilibrium results in an student policy capable of solving all possible solvable levels, then regret is maximised by generating solvable levels. Therefore, a good regret approximation metric should not result in high scores for unsolvable levels. Both PVL and MaxMC implicitly score low for unsolvable levels. For the Nash equilibrium result to hold, the reward conditions \cite{dennis2020emergent} necessitate that the maximum achievable return for an unsolvable level $F_{max}$ must not exceed the minimum return achievable in a solved level $S_{min}$. In the case of MaxMC, a higher $R_{max}$ can be achieved if the level is solvable, and so solvable levels will generally score higher. For PVL, higher returns will tend to result in higher advantage, and therefore a higher score, so solvable levels that can achieve higher returns will score high.

The issue with this implicit bias towards solvable levels is that in practice, this manifests as a bias towards levels with a high success rate \cite{rutherford2024no}, i.e. the current policy solves the level with a high probability. Therefore, these metrics generally do not produce sufficiently challenging levels for training. SFL \cite{rutherford2024no} has shown that training using levels with approximately $50\%$ success rate results in strong training performance. However, determining the exact success rate for a given level requires substantially more environment rollouts than necessary for metrics such as PVL or MaxMC.

While MNA is capable of identifying challenging levels with significantly fewer rollouts than directly scoring based on success rate, over-approximations of state value $\hat{V}(s)$ may result in high scores for unsolvable levels. In order to compensate for this, we introduce an explicit penalisation for unsolvable levels. As it is often intractable to determine exact solvability of levels in complex environments, we instead introduce approximate unsolvablilty, where we define a level as approximately unsolvable if it has never been solved, e.g. has a success rate of $0\%$. Therefore, if a level is approximately unsolvable, we set the score for the level to zero. As such, our final proposed regret approximation is
\begin{equation}
    \text{MNA} = \left(\dfrac{1}{T} \sum_{t=0}^T\hat{G}_t^{\lambda} \right) \cdot \hat{C}
\end{equation}
where $\hat{C}$ is $0$ if the level is approximately unsolvable and $1$ otherwise.

\section{Experimental Setup}
For this work, we examine the standard minigrid environment used in previous UED work \cite{dennis2020emergent, jiang2021replay, parker2022evolving}, as well as evaluating UED performance on the modified minigrid with the addition of a key and locked door. In line with exisiting UED work, our main evaluation metric is zero-shot performance on a set of hand-designed test levels. For the standard minigrid, we use the set of 8 test levels used in previous work \cite{coward2024jaxued, rutherford2024no}. For the modified key minigrid, we modify this set of levels so as to require the agent to unlock the door to reach the goal. Previous work has only examined minigrid when students are trained using 13x13 levels, but in order to scale UED to larger environments, we need to ensure that the student is still capable of learning complex skills, such as unlocking a door, even when trained in larger environments. To assess the student's ability to solve levels that require the door to be unlocked, we evaluate student performance on the existing key minigrid test levels but when trained on levels that are 17x17 and 21x21.

\textbf{Baselines:} In order to assess the effectiveness of both MNA and DEGen, we compare a number of different existing level generation methods and a number of regret approximation metrics. For regret approximation metrics, we compare MNA to the existing metrics, MaxMC and PVL. For assessing these metrics, we use the existing UED methods PLR \cite{jiang2021replay} and ACCEL \cite{parker2022evolving}. We include Domain Randomisation (DR) to show the relative performance of UED compared to a naive, random approach. Additionally, we include SFL \cite{rutherford2024no} for a non-regret based approach. For the standard minigrid environment, we also include the Initial Gen baseline, corresponding to an RL-trained teacher that generates the full environment prior to student rollouts, however we show that this performs substantially worse than all other methods so do not include it in the key minigrid domain. All student agents are trained using PPO \cite{schulman2017proximal}, as well as the teacher agents used in DEGen and Initial Gen. Detailed training hyperparameters for all domains and UED methods are found in Appendix \ref{app:Hyperparameters}.

\section{Results}

\subsection{Minigrid}

\begin{figure}[ht]
  \centering
  \includegraphics[width=0.8\textwidth]{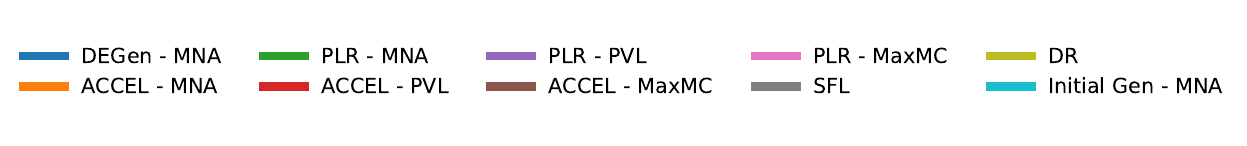}\\[-1.5em]
  
  \subfigure[Mean Return]{
    \includegraphics[width=0.48\textwidth]{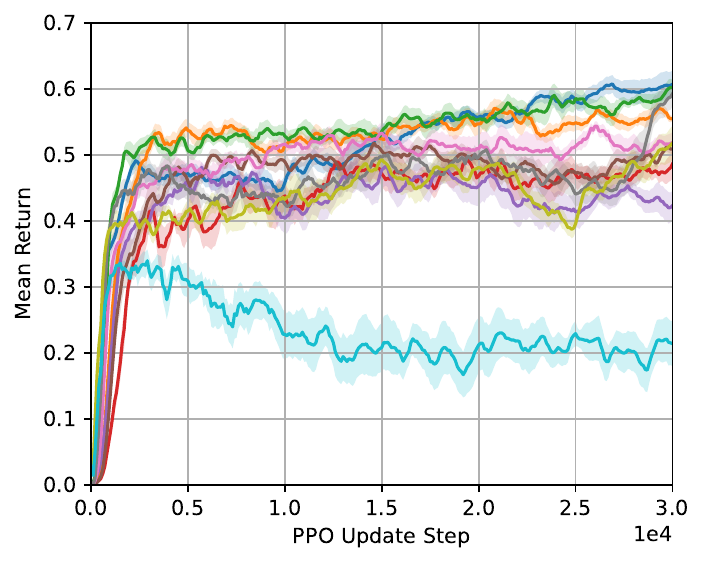}
    \label{fig:Minigrid_Mean_Return}
  }
  \subfigure[Mean Solve Rate]{
    \includegraphics[width=0.48\textwidth]{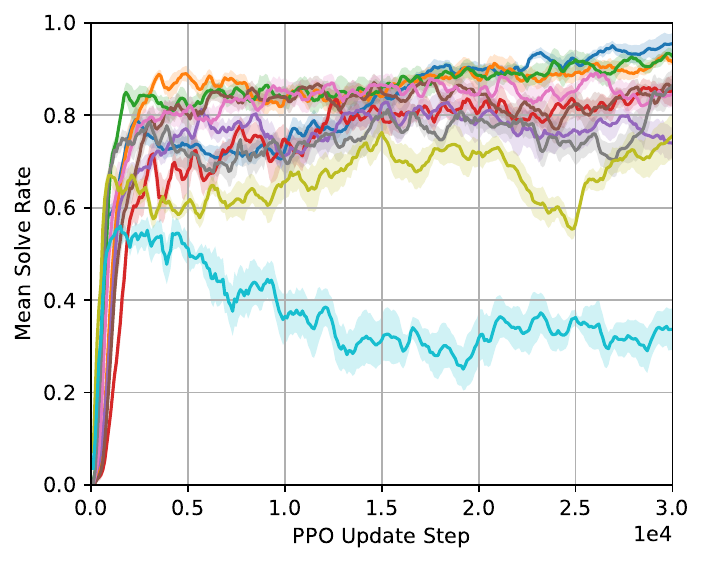}
    \label{fig:Minigrid_Mean_Solve_Rate}
  }
  \caption{Minigrid zero-shot performance on hand-designed test set, showing mean and standard error across 8 runs.}
  \label{fig:Minigrid_Results}
\end{figure}

Figure \ref{fig:Minigrid_Results} illustrates the performance of MNA and DEGen compared to existing baselines. From these plots, it is clear that Intial Gen performs substantially worse than all other methods, whereas DEGen performs comparably to existing replay-based methods. This substantial performance deficit can likely be attributed to a lack of diversity in the generated levels, see Appendix \ref{app:Training_Level_Examples}. We see a far greater diversity in the levels generated via DEGen, and this, along with the reduced credit assignment challenge, allows for DEGen to substantially outperform Initial Gen, despite the former also relying on an RL-trained teacher. We also see that for both PLR and ACCEL, MNA outperforms the existing MaxMC and PVL regret approximations. However, Figure \ref{fig:Minigrid_Results} also shows that in this setting, early in training, DEGen is outperformed by replay-based methods using MNA. This suggests that, whilst the final DEGen performance is equal or greater than the performance of replay-based methods, the additional challenge of learning the generator does impact initial performance.

\begin{figure}[ht]
  \centering
  \includegraphics[width=0.8\textwidth]{graphics/MINIGRID_LEGEND.pdf}\\[-1.5em]
  
  \subfigure[Mean Return]{
    \includegraphics[width=0.48\textwidth]{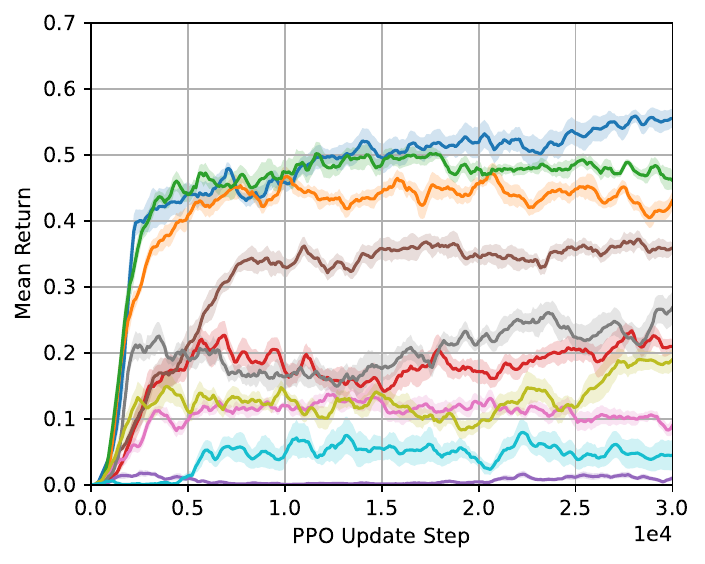}
    \label{fig:Key_Minigrid_Mean_Return}
  }
  \subfigure[Mean Solve Rate]{
    \includegraphics[width=0.48\textwidth]{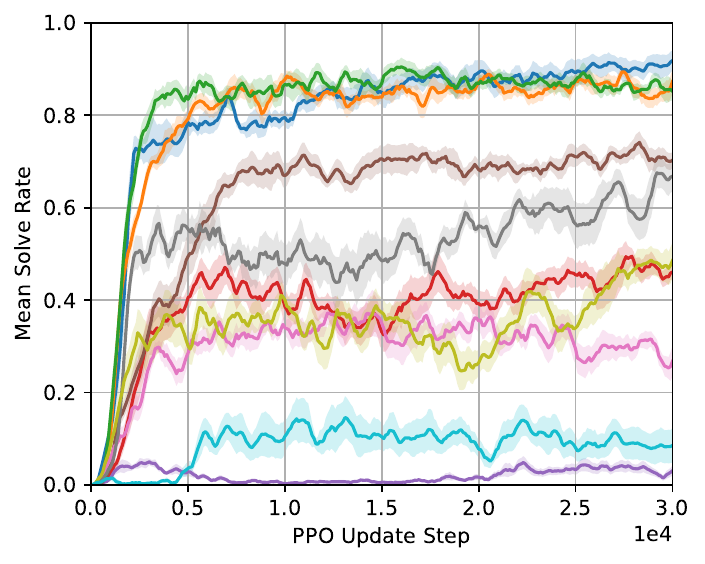}
    \label{fig:Key_Minigrid_Mean_Solve_Rate}
  }
  \caption{Minigrid with key and locked door zero-shot performance on hand-designed test set, trained on 13x13 training levels, showing mean and standard error across 8 runs.}
  \label{fig:Key_Minigrid_Results}
\end{figure}

\begin{figure}[ht]
  \centering
  \includegraphics[width=0.5\textwidth]{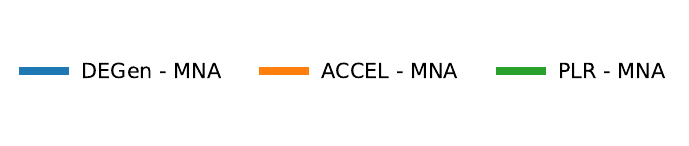}\\[-2em]
  
  \subfigure[17x17]{
    \includegraphics[width=0.48\textwidth]{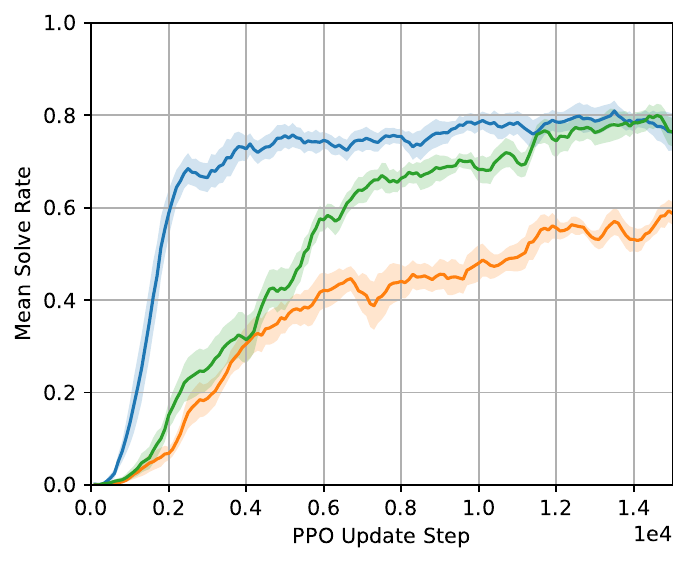}
    \label{fig:Key_Minigrid_Mean_Solve_Rate_17}
  }
  \subfigure[21x21]{
    \includegraphics[width=0.48\textwidth]{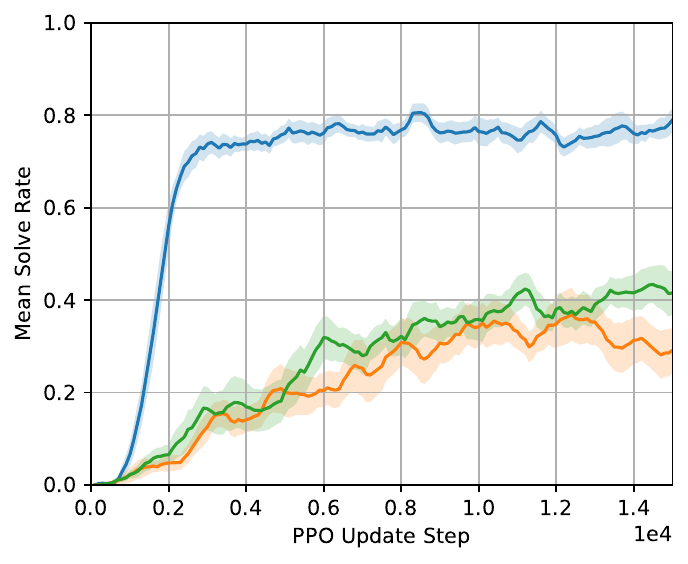}
    \label{fig:Key_Minigrid_Mean_Mean_Solve_Rate_21}
  }
  \caption{Minigrid with key and locked door zero-shot performance on hand-designed test set, trained on larger training levels, showing mean and standard error across 8 runs.}
  \label{fig:Large_Key_Minigrid_Results}
\end{figure}

\subsection{Key Minigrid}

Figure \ref{fig:Key_Minigrid_Results} compares the performance of the various tested UED methods in the key minigrid domain. Due to the increased challenge of level generation with the addition of a key and locked door, there is far more variance in the relative performance of each of the methods. In this domain, we see DEGen outperform all existing baselines, as well as the MNA-based replay methods. We see that PLR using either MaxMC or PVL performs extremely poorly, whereas PLR using MNA, which is much more capable of identifying challenging levels, is the best performing baseline. This highlights the improved regret approximation of MNA compared to existing metrics. We also note that SFL performs poorly in this key-minigrid setting. Learnability only considers the final outcome of an episode, rather than the full student trajectory, and this result suggests that this may be insufficient in domains where additional environment features result in a more complex subset of levels. 

\subsection{Increased Environment Size for Key Minigrid}
Up to this point, we have examined only the relatively small 13x13 environment setting. However, as outlined above, with the aim of scaling UED to larger, more complex environments, we examine the performance of current UED methods when simply scaling up a version of minigrid to sizes larger than 13x13. Figure \ref{fig:Large_Key_Minigrid_Results} shows the performance of DEGen, PLR and ACCEL using MNA when training levels are 17x17 and 21x21. Whilst there is a performance drop compared to the 13x13 levels, as the size of the environment grows, DEGen substantially outperforms existing methods. This result highlights that, with any domains with additional environment features such as the key and door that can potentially add additional complexity to policy learning, as the environment size grows, it becomes substantially harder to sample useful levels where these features are necessary. Therefore it is necessary to use a trained generator and DEGen is able to overcome the credit assignment challenges present in previous methods for training learnt generators.

\section{Discussion}

It is clear that from these results that DEGen does substantially outperform existing baselines in both the minigrid, and key minigrid environments, and this performance improvement is increased as the environment size increases. Scaling environment size is likely a necessity if future work aims to move UED from the current set of toy small-scale environments used, to more useful real-world domains, and this work shows that replay-based methods perform poorly with relatively small increases to environment size. We have demonstrated that DEGen is an effective method of addressing these issues, both compared to generating the full level prior to student rollouts, and replay-based methods.  

\section{Limitations}
\label{sec:Limitations}

Whilst the results we have presented in this paper do show strong performance from DEGen compared to existing baselines, the domains presented in this paper present relatively simple mappings between the level representation and the agent's current observation. In more complex domains, such as 3D Games \cite{harmer2018imitation, ferdous2022towards, hafner2025training}, or real-world robotics applications \cite{nguyen2019review, zhu2021deep, han2023survey}, it will be more complex to determine how specific environment parameters affect what the agent is currently observing. In order for UED to bridge the gap from the current set of game domains to real-world applications, it would be necessary for DEGen or DEGen-like methods to address this limitation. We believe that World Models \cite{ha2018world} represent a promising direction for future research to address this. In its current form, DEGen relies on both the agent and the generator interacting with a fixed environment, where the environment has an explicit mapping between the agent's observation and the level parameters, and the generator is only able to generate the level where the agent has observed. However, a world model guided by a regret approximation such as MNA would represent a generator that could directly generate observations for the agent. Rather than relying on explicit mapping between level and observation, this mapping could be learnt with environment data when training the world model. Whilst the training of a world model would add additional computational cost to the training process, it would enable DEGen methodology to be applied to substantially more complex environments. With the advent of highly general world models such as the Genie series of world models \cite{bruce2024genie}, this could represent a path to training highly general policies that are effective in a wide variety of applications.

Additionally, in line with previous work \cite{jiang2021replay, parker2022evolving, chung2024adversarial}, we have compared the relative performance of UED methods based on the number of student PPO update steps. However, training the DEGen teacher agent adds a high computational cost to the training loop, and so training using DEGen takes approximately four times as long as training using methods such as PLR and ACCEL. Full details on compute time and experiment specification can be found in Appendix \ref{app:Experimental_Setup}. Therefore, replay-based methods may be preferable for small environment sizes where similar performance is achieved. However again, it is clear than as environment size grows, the maximum performance achieved by DEGen exceeds replay-based methods, and so the additional time cost is justified, given the performance gains.  

\section{Conclusion}
In this paper, we introduce a new level generation method, Dynamic Environment Generation for UED, and a new regret approximation metric, Maximised Negative Advantage. We outline how current UED methods fail as training environment size increases, and show that DEGen is capable of mitigating the issues associated both with these larger environments, and with RL-based level generation. We show that the use of MNA enables DEGen to outperform existing baselines, whilst also showing that the use of MNA consistently improves the performance of existing UED methods. These performance improvements are most evident in the more complex key minigird domain.  We believe there is significant potential for future UED research to address larger and more complex environments, and that approaches based on MNA and DEGen provide a promising foundation for this advancement.

\section*{Acknowledgements}
This work was supported by the EPSRC Centre for Doctoral Training in Autonomous Intelligent
Machines and Systems [EP/S024050/1]. Lacerda and Hawes have received EPSRC funding via the “From Sensing to Collaboration” programme grant [EP/V000748/1]. 

\newpage

\medskip

\small

\bibliographystyle{plain}
\bibliography{bibliography}

\appendix

\newpage

\section{Related Work}

This research presents a new method for Unsupervised Environment Design (UED) \cite{dennis2020emergent}, an area of research focused on improving the generalisation of policies trained using reinforcement learning and enabling open-ended learning. Domain Randomisation (DR) can be categorised as the most basic form of UED, where environment parameters are simply randomised naively for training. DR has shown to be effective in applications such transferring policies from simulations to real world robotics deployments \cite{tobin2017domain, peng2018sim, andrychowicz2020learning}. Prioritised Level Replay (PLR) \cite{jiang2021prioritized, jiang2021replay} augments domain randomisation by maintaining a replay buffer of levels that have been scored as effective for training. PLR has been employed in a diverse set of domains, such as in training World Models \cite{rigter2024reward} or meta-reinforcement learning \cite{jackson2023discovering}. Additional work has examined dealing with distribution shift between training and deployment \cite{jiang2022grounding}. Whilst PLR relies on replaying levels from a fixed random generator, other work has focused on generating new levels through evolving prior levels \cite{parker2022evolving}. POET \cite{wang2019paired, wang2020enhanced} examines co-evolving both the environment levels, but also a population of agents playing those levels. More similarly to DEGen, other existing work focuses on explicitly learning a level generator \cite{dennis2020emergent, azad2023clutr, mediratta2023stabilizing, chung2024adversarial}.

PAIRED \cite{dennis2020emergent} introduced the formalisation for UED, framing the level design problem as a minimax regret game between the student and teacher. PAIRED relies on the performance difference between a protagonist and antagonist agent to score level suitability, and uses these scores to train a level generator. CLUTR \cite{azad2023clutr} improves on PAIRED by learning a latent representation of levels to reduce the challenge of learning the generator. ADD \cite{chung2024adversarial} has used a diffusion model to generate levels instead. More recent work has examined using non-regret based scoring of levels \cite{tzannetos2023proximal, tzannetos2024proximal} such as learnability \cite{rutherford2024no, monette2025optimisation, foster2025learning}. 

UED can be seen as a form of Automatic Curriculum Learning (ACL) \cite{portelas2020automatic, florensa2018automatic}, where ACL aims to provide an automatic curriculum to enable learning of increasingly challenging tasks. Unlike UED, ACL often relies on specific knowledge of the target task \cite{fang2021adaptive, matiisen2019teacher}. 

This work also relates to the field of Procedural Content Generation (PCG) \cite{liapis2020ten, risi2020increasing}. PCG has focused on level design for games, such as with terrain generation \cite{smelik2010integrating}, task design \cite{doran2011prototype, kim2019design} or puzzle setting \cite{khalifa2015automatic, bento2019procedural}. Much of this work has focused on level design for human play, and some of this work relies on specific user input for level generation \cite{liapis2016mixed, campos2017mixed, powley2017automated}. Methods using RL for training level generators \cite{khalifa2020pcgrl, earle2021learning} require hand-designed, environment specific generation rewards, which differs from the domain-agnostic level scoring used in UED.

Unlike previous UED work, DEGen relies on a teacher that generates the environment based on the student's current observations. This shares similarities with existing work in World Models \cite{ha2018world}. Whilst the DEGen teacher is able to control the observations of the student, the environment is designed to explicitly disallow infeasible observations to be generated. World Models are instead trained on example trajectories, and the feasibility of generated observations is implicitly learnt. These learnt models are capable of training agents without access to the real environment \cite{robine2023transformer, micheli2023transformers, hafner2025training}. Large open-ended world models \cite{yang2023learning, hu2023gaia, bruce2024genie} may potentially enable the training of highly generally capable agents . Recent work has examined applying UED methods, specifically PLR, to training agents in world models \cite{guzel2025imagined}.

\newpage

\section{Detailed Experimental Setup} \label{app:Experimental_Setup}
\subsection{Environment Details}
\subsubsection{Minigrid}

We use the standard minigrid implementation from exisiting UED work \cite{dennis2020emergent, jiang2021replay, parker2022evolving, rutherford2024no, chung2024adversarial}. Examples of levels are shown in Figure \ref{fig:Minigrid_Eval_Levels}. 

\textbf{Observations:} The agent, depicted in red, observes a 5x5 square in front of it. It also has access to its absolute direction, e.g. whether it is facing \textit{North}, \textit{South}, \textit{East} or \textit{West}.

\begin{figure}[ht]
  \centering
  \subfigure[Environment State]{
    \includegraphics[width=0.25\textwidth]{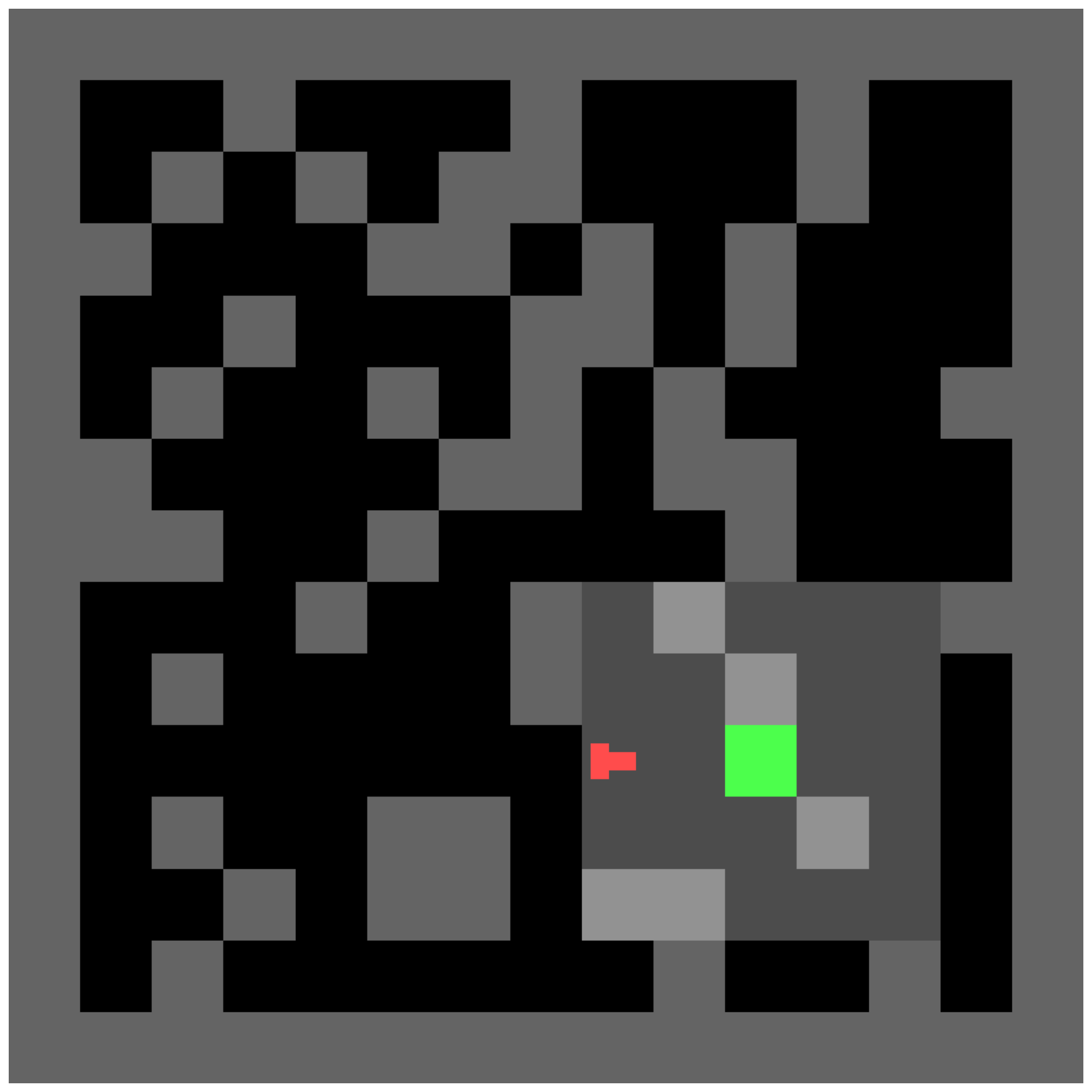}
    \label{fig:Example_Level}
  }
  \hspace{1cm}
  \subfigure[Agent Observation]{
    \includegraphics[width=0.25\textwidth]{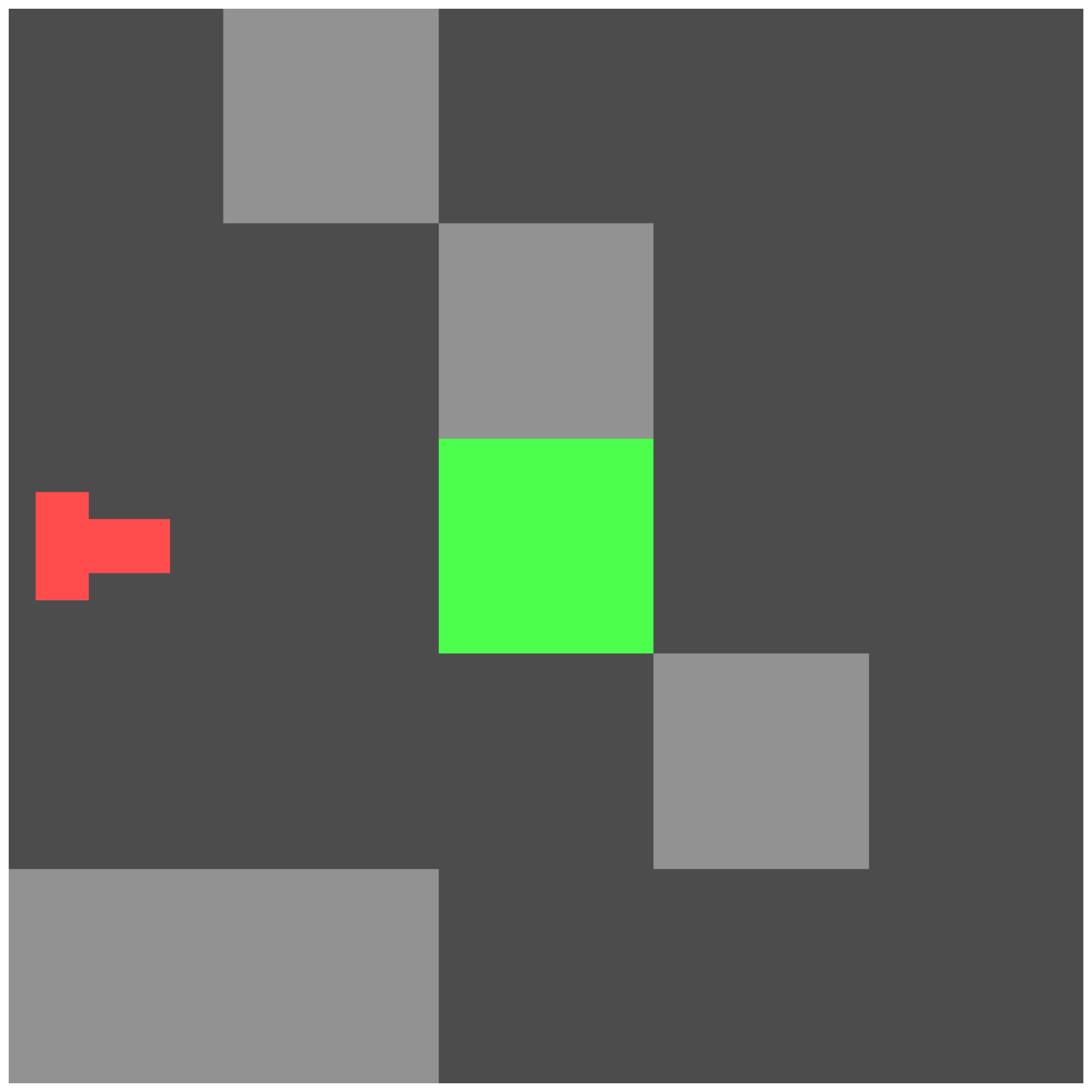}
    \label{fig:Example_Level_Observation}
  }
  \caption{Minigrid Agent Observations}
  \label{fig:Example_Observation}
\end{figure}

\textbf{Rewards:} The agent receives reward $r_g$ if it reaches the goal. The size of this reward is determined by the number of steps taken to reach the goal $T$, and the parameter $T_{\max}$, specifying the maximum number of steps before an episode terminates. 
\begin{equation}
    r_g = 1 - 0.9\left(\dfrac{T}{T_{\max}}\right)
\end{equation}

\textbf{Actions:} At each step, the agent can either move \textit{forward}, turn \textit{left} or turn \textit{right}.

\subsubsection{Key Minigrid}
The key minigrid implementation is identical to the standard minigrid level, except for the the addition of a key and locked door.

\textbf{Observations:} The agent is additionally able to observe whether it has collected the key.

\textbf{Rewards:} The reward remains identical to the standard minigrid, with reward only being received on reaching the goal. Note that there are no specific rewards associated with collecting the key or unlocking the door.

\textbf{Actions:} The agent has an additional \textit{use} action. The agent will pick up the key if it reaches the grid square the key is on. In order to unlock the door, the agent must select the \textit{use} action when it has the key and the door is directly ahead of the agent.

\begin{figure}[ht]
  \centering
  \subfigure[Locked Door]{
    \includegraphics[width=0.23\textwidth]{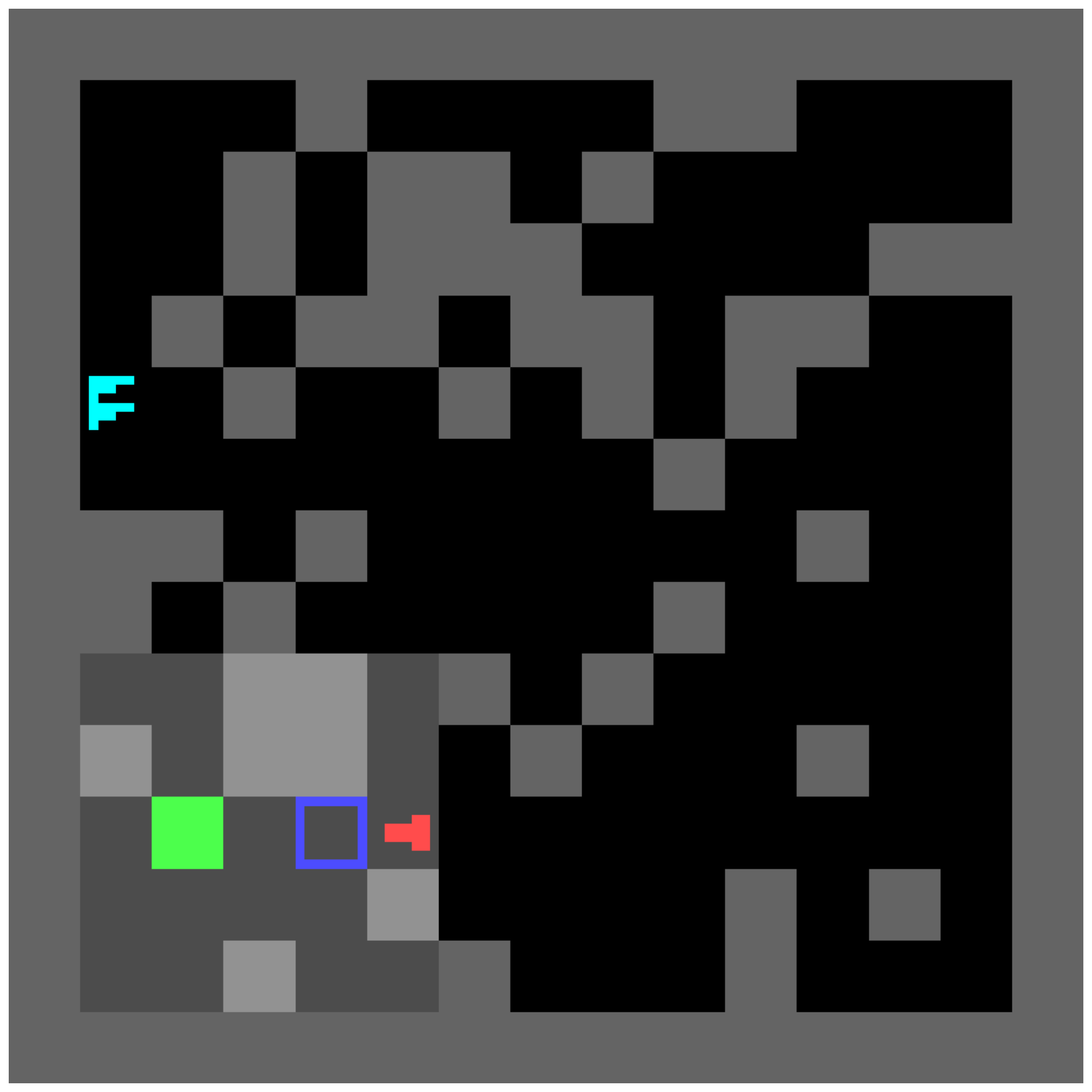}
    \label{fig:Example_Unlock_1}
  }
  \subfigure[\textit{use} action]{
    \includegraphics[width=0.23\textwidth]{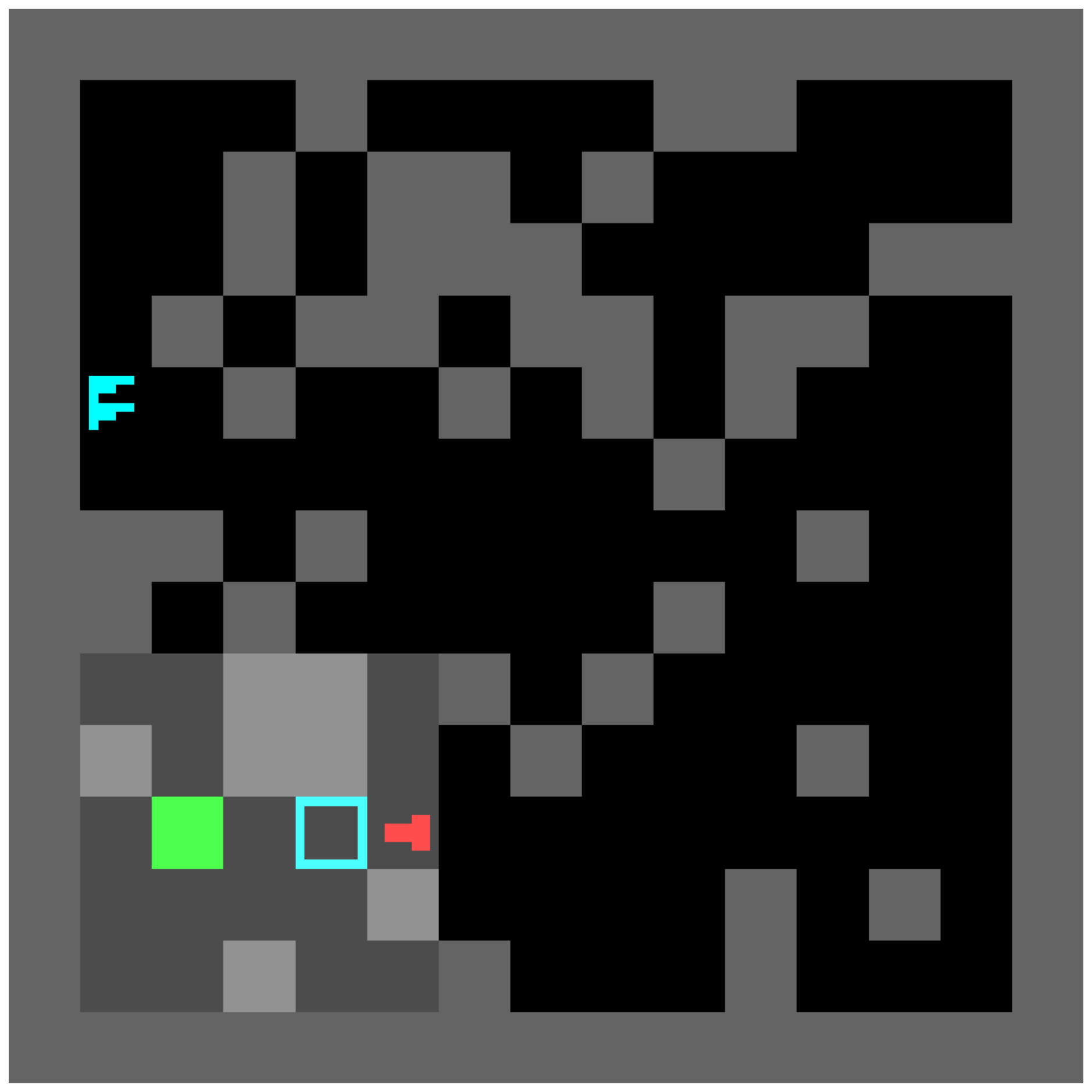}
    \label{fig:Example_Unlock_2}
  }
  \subfigure[Unlocked Door]{
    \includegraphics[width=0.23\textwidth]{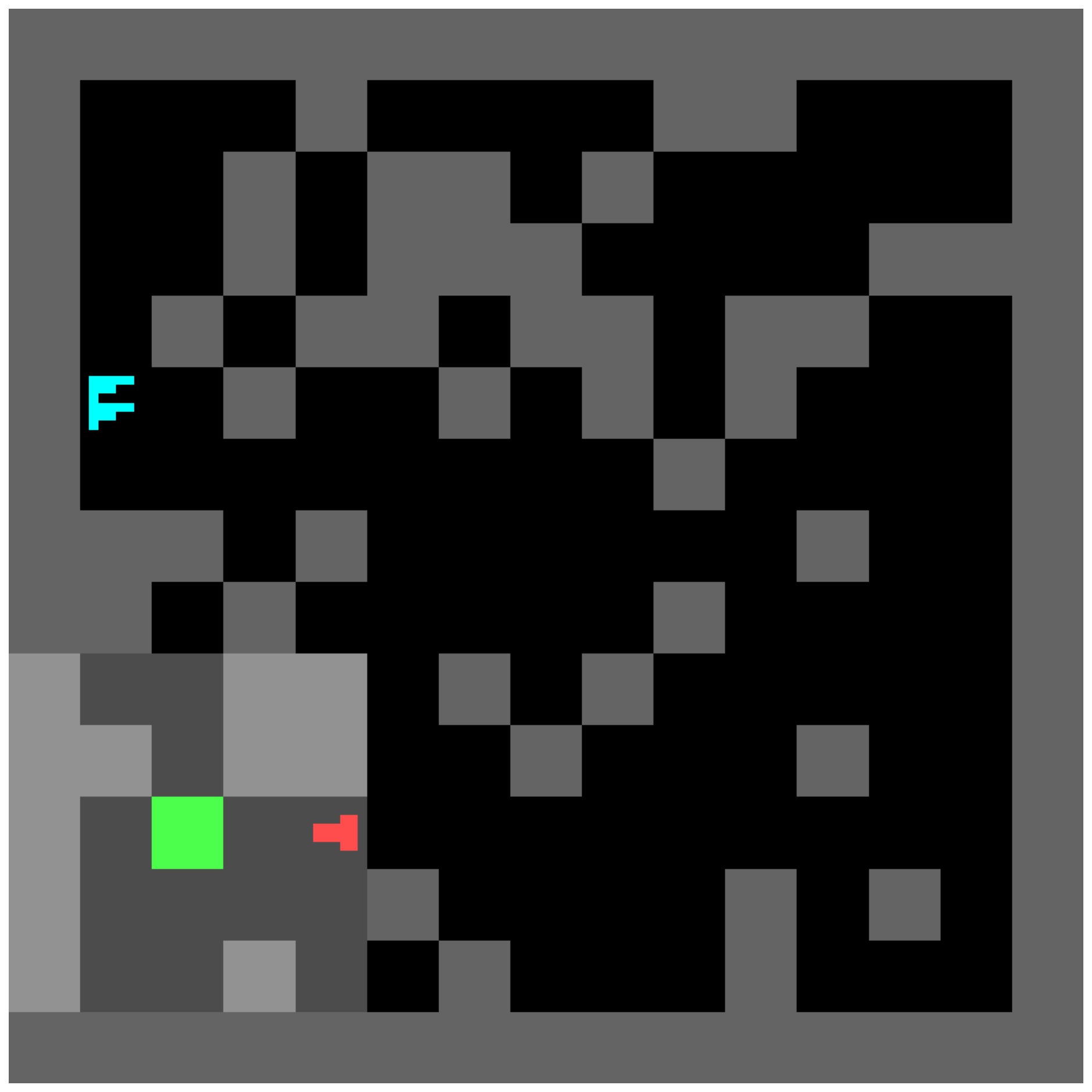}
    \label{fig:Example_Unlock_3}
  }
  \caption{Key Minigrid Door Unlocking}
  \label{fig:Example_Unlock}
\end{figure}

\newpage
\subsubsection{DEGen Teacher}
\textbf{Observations:} The teacher's observations are an augmented version of the student agent. Similarly to the student, the teacher observes a 5x5 square in front of the student. However, this is augmented by an overlayed 5x5 square that indicates which squares have yet to be generated.

\begin{figure}[ht]
  \centering
  \subfigure[Environment State]{
    \includegraphics[width=0.25\textwidth]{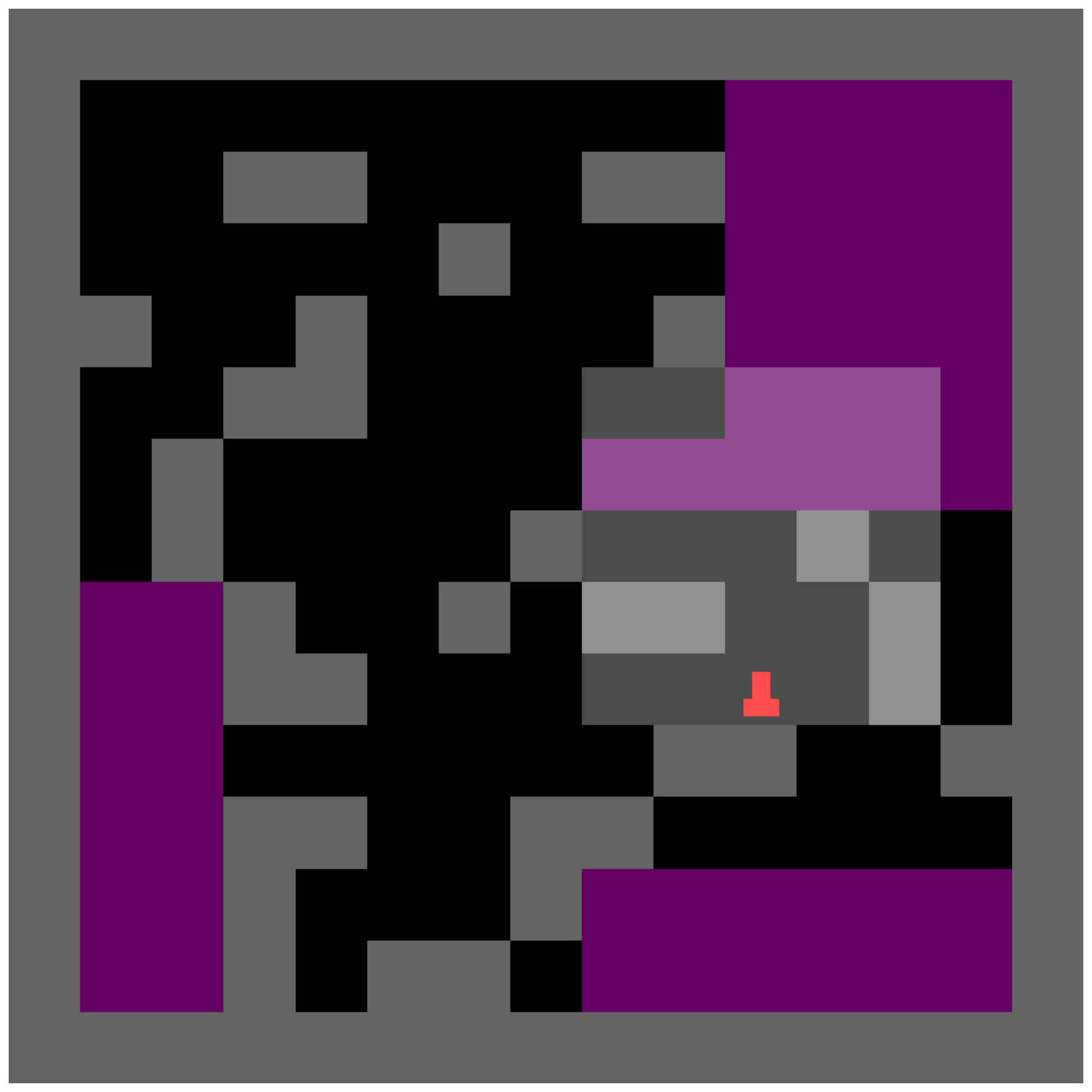}
    \label{fig:DEGen_Teacher_Env_State}
  }
  \hspace{1cm}
  \subfigure[Teacher Observation]{
    \includegraphics[width=0.25\textwidth]{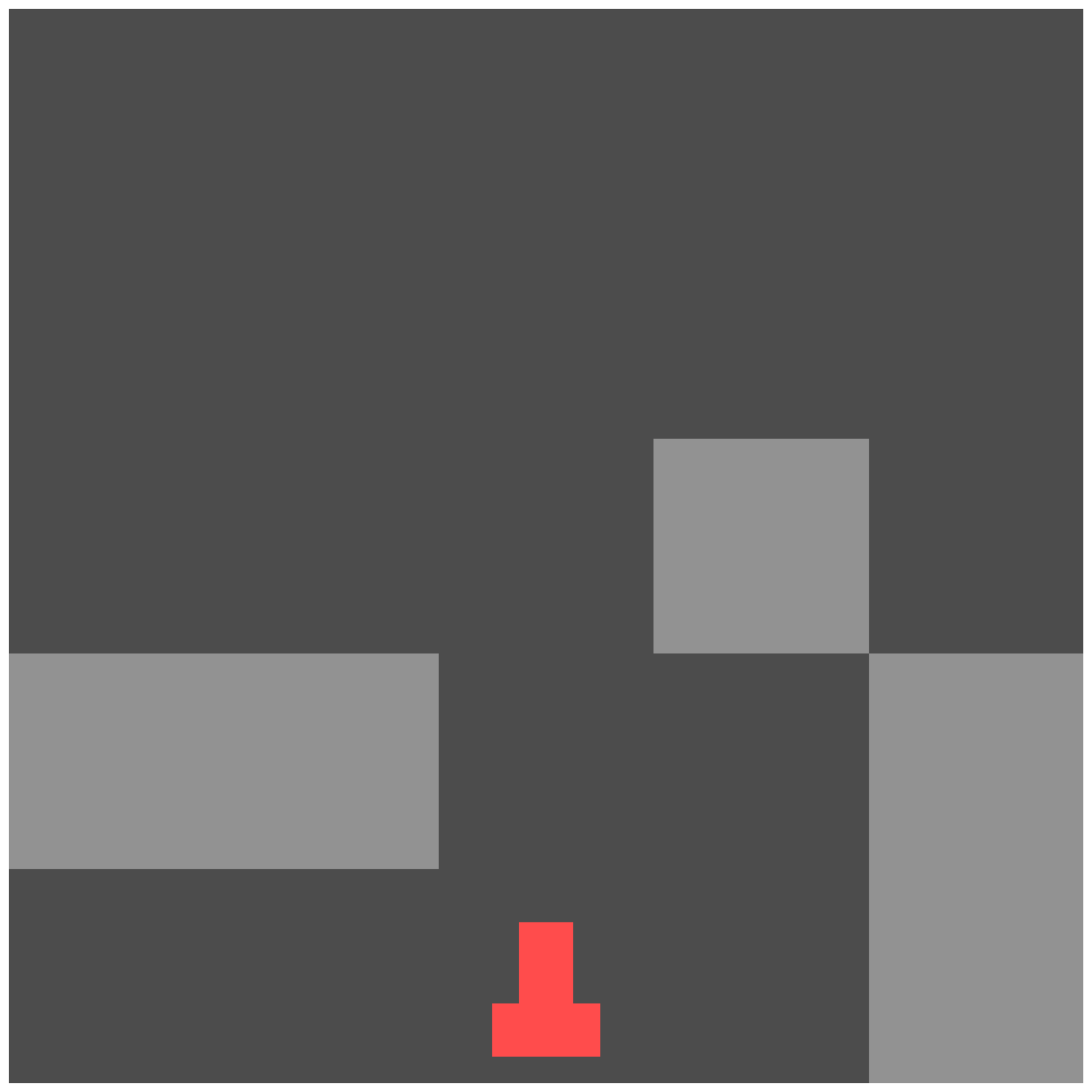}
    \includegraphics[width=0.25\textwidth]{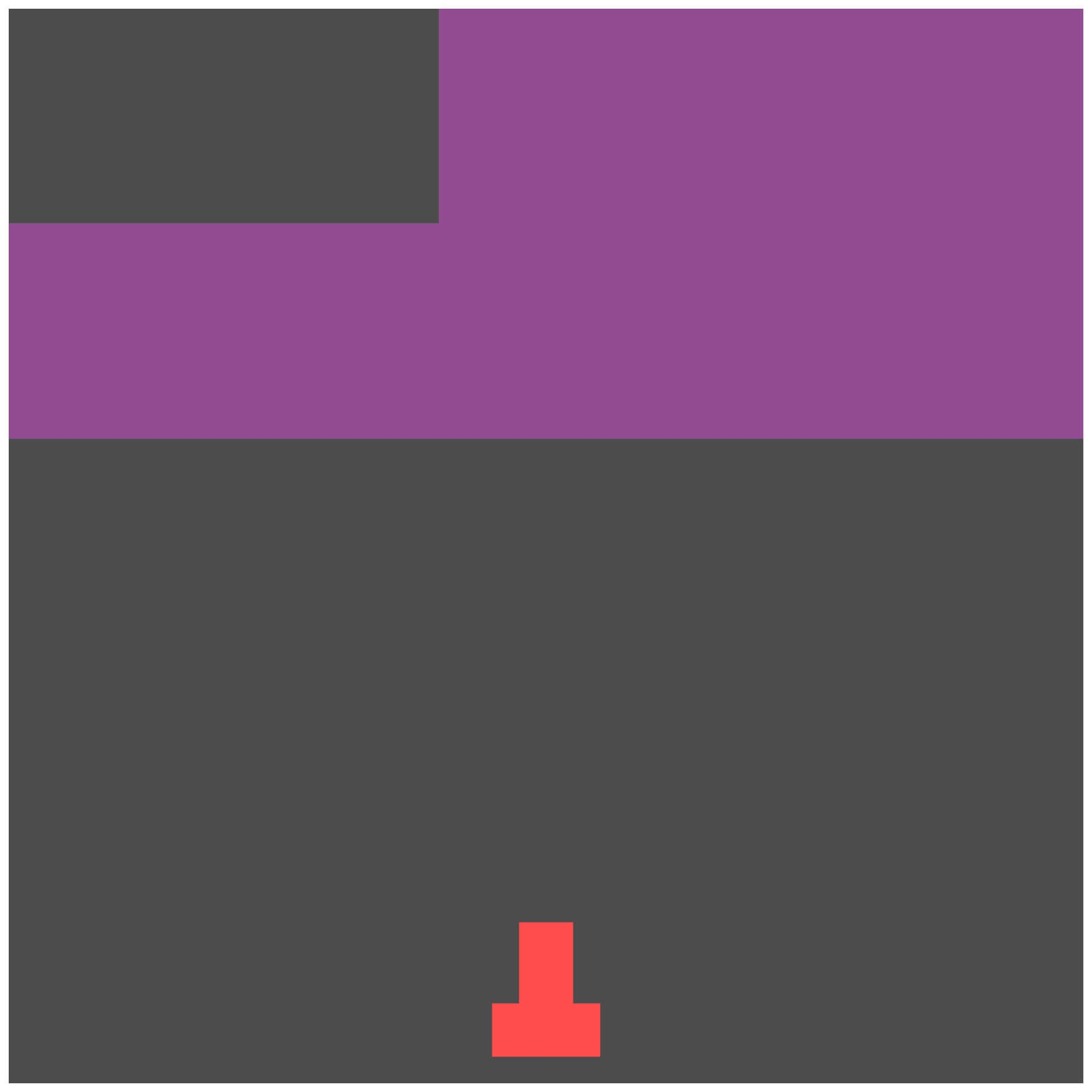}
    \label{fig:DEGen_Teacher_Obseration}
  }
  \caption{DEGen Teacher Observations}
  \label{fig:DEGen_Teacher_Observations}
\end{figure}

\textbf{Actions:} At each step, the teacher is able to fill any ungenerated grid square the student is currently observing. The teacher action $a$ is composed of two sub-actions $(a_1, a_2)$ where $a_1$ selects which cell will be filled, and $a_2$ selects what the cell will be filled with, e.g. \textit{wall}, \textit{empty}, \textit{goal}, \textit{key}, \textit{door}. We use action masking to ensure that only previously ungenerated cells are filled, as well as enforce that only one \textit{goal}, \textit{key} and \textit{door} can be placed for each level.

\textbf{Rewards:} The teacher rewards are determined by the chosen regret approximation. These dense rewards are calculated using Equation \ref{eqn:Dense_Teacher_Rewards}.

\textbf{Dynamics}: As the environment is rolled out, at each step, either the teacher or the student will act. If there are ungenerated cells in the agent's observation, the teacher must fill all ungenerated cells currently observed before the student is able to take another action. 

As both the student and teacher have partial observability, both policy networks include an LSTM layer, so actions are conditioned on all previous observations. The teacher is conditioned on all previous observations, whereas the student is conditioned only on  the fully-generated observations.

\subsubsection*{KL Regularisation}
Prior work \cite{mediratta2023stabilizing} has shown that entropy regularisation is necessary when using reinforcement learning to train a teacher for UED. The policy gradient loss is augmented with an entropy loss
\begin{equation}
    L_\text{entropy} = \mathbb{E}_{s \sim \pi} \biggl[ -H \Bigl( \pi(\cdot|s) \Bigr) \biggr].
\end{equation}
Entropy regularisation biases the policy towards a more uniform distribution of action probabilities. However, as there can only be a maximum of a single \textit{goal}, \textit{key} and \textit{door} per level, the distribution of cell objects will be highly non-uniform, as the vast majority of cells will be either \textit{wall} or \textit{empty}. Therefore rather than including an entropy loss for the $a_2$ sub-action, we instead add KL regularisation with a fixed categorical distribution $q(a_2)$, defined with the parameter $p_g$, such that
\begin{equation}
    q(a_2) \;=\; \mathrm{Cat}(\boldsymbol{\alpha}), 
\qquad 
\boldsymbol{\alpha} = (p_w,\;p_w,\;p_g,\;p_g,\;p_g),
\qquad 
p_w = \dfrac{1-3p_g}{2}
\end{equation}
and 
\begin{equation}
    L_{emtopy\; + \: KL} = \mathbb{E}_{s \sim \pi} \biggl[ -H \Bigl( \pi_{a_1} (\cdot|s) \Bigr) + D_{KL} \Bigl( \pi_{a_2} (\cdot|s) \Big| \Big| q(\cdot) \Bigr )\biggr].
\end{equation}
For all experiments, we used $p_g=0.01$.

\newpage

\subsection{Hyperparameters} \label{app:Hyperparameters}
Full code and instructions on how to run can be found at \href{https://github.com/HarryMJMead/Dynamic-Environment-Generation-for-UED}{\texttt{https://github.com/HarryMJMead/Dynamic-Environment-Generation-for-UED}}. All existing methods were trained using implementations based on JaxUED \cite{coward2024jaxued}, available at \href{https://github.com/DramaCow/jaxued}{\texttt{https://github.com/DramaCow/jaxued}}, and SFL \cite{rutherford2024no}, available at \href{https://github.com/amacrutherford/sampling-for-learnability}{\texttt{https://github.com/amacrutherford/sampling-for-learnability}}. Learning hyperparameters are shown in Table \ref{tab:Learning_Hyperparameters} and the replay UED hyperparameters are shown in Table \ref{tab:UED_Hyperparameters}.

\begin{table}[H]
\centering
\caption{Learning Hyperparameters.}
\label{tab:Learning_Hyperparameters}
\begin{tabular}{llll}
\toprule
\textbf{Parameter} & \textbf{Minigrid} & \textbf{Key Minigrid 13x13} & \textbf{17x17 and 21x21}  \\
\midrule
\multicolumn{3}{l}{\textbf{Student PPO}} \\
Number of Updates & 30000 & & 15000\\
$\gamma$ & 0.995 & & \\
$\lambda_{\text{GAE}}$ & 0.95 & & \\
PPO number of steps & 512 & & \\
PPO epochs & 4 & & \\
PPO minibatches per epoch & 4 & & \\
PPO clip range & 0.04 & & \\
PPO \# parallel environments & 256 & & \\
Adam learning rate & 5e-4 & & 2.4e-4 \\
Anneal LR & yes &  & no \\
Adam $\epsilon$ & 1e-5 & & \\
PPO max gradient norm & 0.5 & & \\
PPO value clipping & yes & & \\
value loss coefficient & 0.5 & & \\
entropy coefficient & 1e-3 & & \\
Hidden dimension size & 256 & & \\
\midrule
\multicolumn{3}{l}{\textbf{Teacher PPO}} \\
$\gamma$ & 0.998 & & 0.99 \\
$\lambda_{\text{GAE}}$ & 0.95 & & \\
PPO epochs & 4 & & \\
PPO minibatches per epoch & 4 & & \\
PPO clip range & 0.2 & & \\
Adam learning rate & 1e-3 & & \\
Anneal LR & yes &  & no \\
Adam $\epsilon$ & 1e-5 &  & \\
PPO max gradient norm & 0.5 & & \\
PPO value clipping & yes & & \\
value loss coefficient & 0.5 & & \\
entropy coefficient & 5e-2 & & \\
Hidden dimension size & 256 & & \\
Num Teacher Steps \textit{(Initial Gen)} & 60 & & \\
\bottomrule
\end{tabular}
\end{table}

\newpage

\begin{table}[H]
\centering
\caption{UED Hyperparameters.}
\label{tab:UED_Hyperparameters}
\begin{tabular}{lll}
\toprule
\textbf{Parameter} & \textbf{Minigrid} & \textbf{Key Minigrid} \\
\midrule
\multicolumn{3}{l}{\textbf{PLR}} \\
Replay rate, $p$ & 0.5 &  \\
Buffer size, $K$ & 8000 &  \\
Prioritisation & Rank & \\
Temperature, $\beta$ & 1.0 & \\
staleness coefficient & 0.3 &  \\
\midrule
\multicolumn{3}{l}{\textbf{ACCEL}} \\
Number of Edits & 20 &  \\
Buffer size, $K$ & 8000 &  \\
Prioritisation & Rank &  \\
Temperature, $\beta$ & 1.0 &  \\
\midrule
\multicolumn{3}{l}{\textbf{SFL}} \\
Batch Size $N$ & 25000 &  \\
Rollout Length $L$ & 20000 &  \\
Update Period $T$ & 100 & \\
Buffer Size $K$ & 1000 &  \\
Sample Ratio $\rho$ & 0.5 & \\
\bottomrule
\end{tabular}
\end{table}

\subsection{Compute Details}
For all experiments, each run was on 1 Nvidia A40. We show the mean compute time for both domains and each UED method in Table \ref{tab:Compute_Details}.

\begin{table}[H]
\centering
\caption{Compute Time.}
\label{tab:Compute_Details}
\begin{tabular}{lll}
\toprule
 & \multicolumn{2}{c}{Compute Time (hh:mm)} \\
\textbf{Method} & \textbf{Minigrid} & \textbf{Key Minigrid} \\
\midrule
DR & 11:16 ± 00:02 & 12:01 ± 00:02 \\
Initial Gen & 13:39 ± 00:01 & 13:49 ± 00:02 \\
PAIRED & 23:32 ± 00:02 & 23:33 ± 00:01 \\
SFL & 10:53 ± 00:01 & 10:48 ± 00:02 \\
PLR & 06:53 ± 00:02 & 06:53 ± 00:01 \\
ACCEL & 05:43 ± 00:02 & 05:45 ± 00:01 \\
DEGen & 25:49 ± 00:00 & 25:53 ± 00:01 \\

\bottomrule
\end{tabular}
\end{table}

\newpage

\subsection{Zero-shot Transfer Levels}
\begin{figure}[H]
\vskip 0.2in
\begin{center}
\centerline{\includegraphics[width=0.8\textwidth]{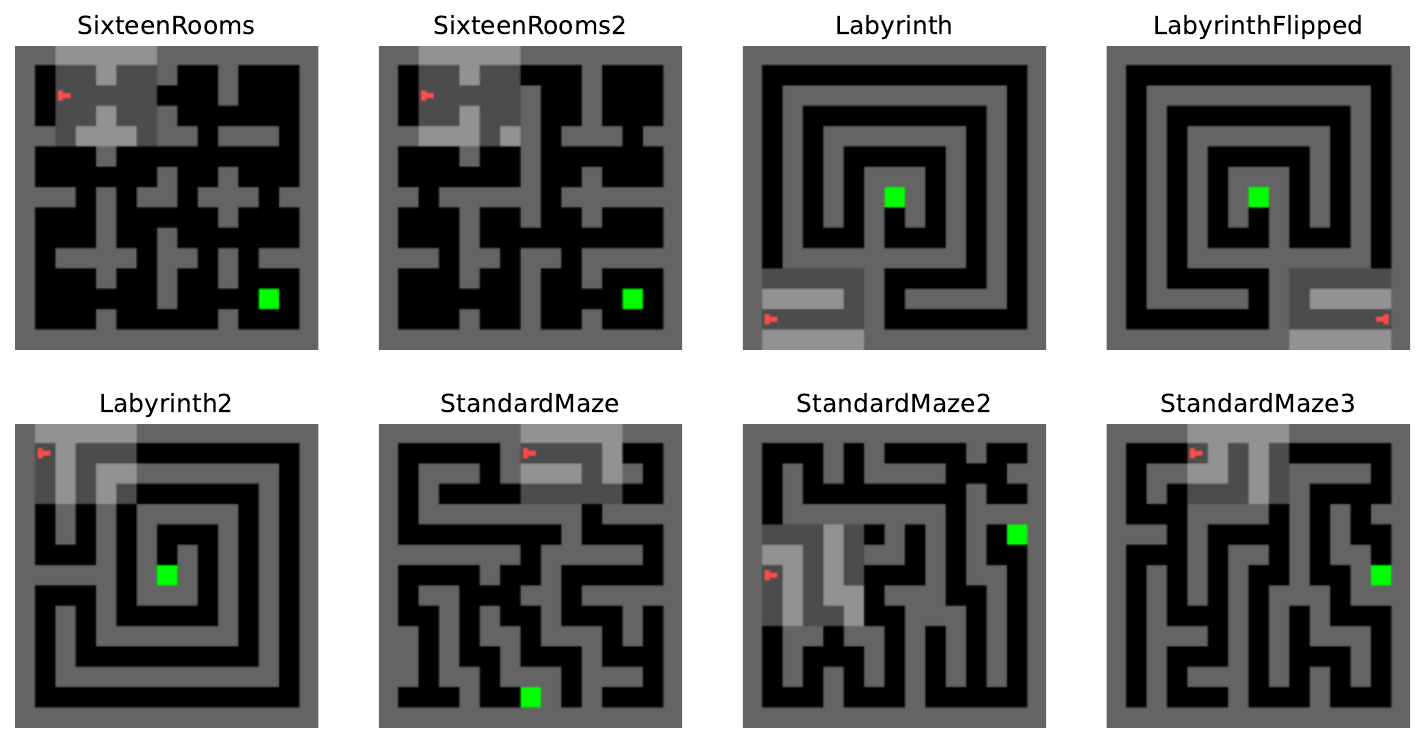}}
\caption{Hand designed evaluation levels for minigird}
\label{fig:Minigrid_Eval_Levels}
\end{center}
\vskip -0.2in
\end{figure}

\begin{figure}[H]
\vskip 0.2in
\begin{center}
\centerline{\includegraphics[width=0.8\textwidth]{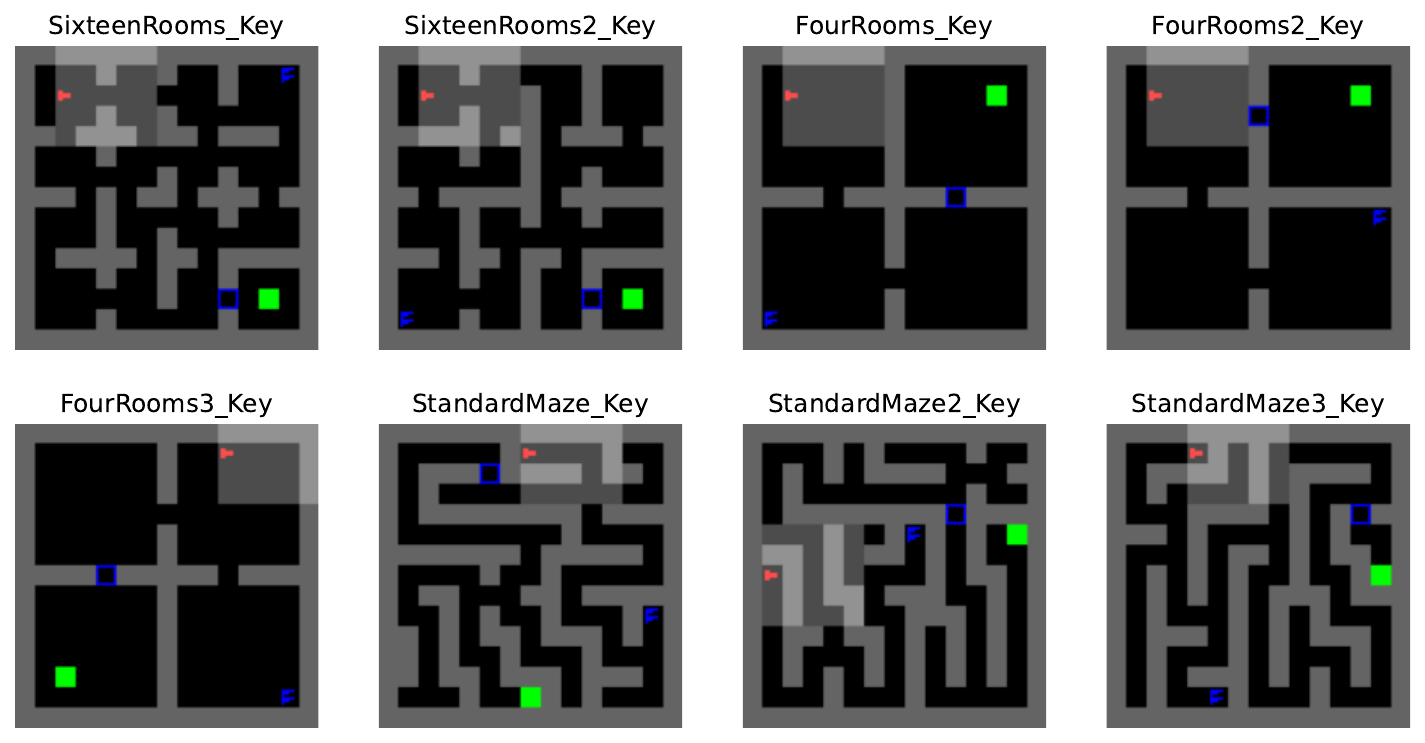}}
\caption{Hand designed evaluation levels for key minigird}
\label{fig:Key_Minigrid_Eval_Levels}
\end{center}
\vskip -0.2in
\end{figure}

Figures \ref{fig:Minigrid_Eval_Levels} and \ref{fig:Key_Minigrid_Eval_Levels} show the hand designed levels used for evaluating zero-shot performance. The minigrid levels were taken directly from JaxUED \cite{coward2024jaxued}. The key minigrid levels have been modified so that the student is required to unlock the door to reach the goal. Note that we have chosen to include the FourRooms\_Key levels rather than modified versions of the labyrinth levels, as the key would trivially be on the path to the goal for these labyrinth levels. 

\newpage
\subsection{Tabular Results}
\subsubsection*{Minigrid Results}
\begin{table}[H]
\centering
\caption{Minigrid Solve Rate (1)}
\label{tab:Minigrid_Solve_Rate_1}
\begin{tabular}{llllll}
\toprule
Level & Initial Gen - MNA & DR & SFL & PLR - MaxMC & ACCEL - MaxMC \\
\midrule
SixteenRooms & 0.93 ± 0.06 & \textbf{1.00 ± 0.00} & \textbf{1.00 ± 0.00} & \textbf{1.00 ± 0.00} & \textbf{1.00 ± 0.00} \\
SixteenRooms2 & 0.48 ± 0.14 & \textbf{1.00 ± 0.00} & \textbf{1.00 ± 0.00} & \textbf{1.00 ± 0.00} & \textbf{1.00 ± 0.00} \\
Labyrinth & 0.04 ± 0.04 & 0.51 ± 0.16 & 0.60 ± 0.13 & 0.45 ± 0.12 & 0.69 ± 0.13 \\
LabyrinthFlipped & 0.11 ± 0.10 & 0.38 ± 0.15 & 0.68 ± 0.13 & 0.64 ± 0.12 & 0.50 ± 0.13 \\
Labyrinth2 & 0.05 ± 0.04 & 0.40 ± 0.14 & 0.83 ± 0.12 & 0.75 ± 0.07 & 0.81 ± 0.09 \\
StandardMaze & 0.28 ± 0.11 & 0.99 ± 0.01 & \textbf{1.00 ± 0.00} & 0.98 ± 0.02 & 0.90 ± 0.05 \\
StandardMaze2 & 0.30 ± 0.10 & 0.85 ± 0.06 & 0.95 ± 0.05 & 0.84 ± 0.12 & \textbf{1.00 ± 0.00} \\
StandardMaze3 & 0.56 ± 0.14 & 0.93 ± 0.04 & \textbf{1.00 ± 0.00} & 0.98 ± 0.02 & \textbf{1.00 ± 0.00} \\
\midrule
Mean & 0.34 ± 0.05 & 0.76 ± 0.05 & 0.88 ± 0.04 & 0.83 ± 0.03 & 0.86 ± 0.02 \\
\bottomrule
\end{tabular}
\end{table}

\begin{table}[H]
\centering
\caption{Minigrid Solve Rate (2)}
\label{tab:Minigrid_Solve_Rate_2}
\begin{tabular}{llllll}
\toprule
Level & PLR - PVL & ACCEL - PVL & PLR - MNA & ACCEL - MNA & DEGen - MNA \\
\midrule
SixteenRooms & \textbf{1.00 ± 0.00} & \textbf{1.00 ± 0.00} & \textbf{1.00 ± 0.00} & \textbf{1.00 ± 0.00} & \textbf{1.00 ± 0.00} \\
SixteenRooms2 & 0.95 ± 0.04 & 0.88 ± 0.05 & \textbf{1.00 ± 0.00} & \textbf{1.00 ± 0.00} & \textbf{1.00 ± 0.00} \\
Labyrinth & 0.71 ± 0.09 & 0.86 ± 0.04 & 0.88 ± 0.06 & 0.68 ± 0.09 & \textbf{0.98 ± 0.02} \\
LabyrinthFlipped & 0.63 ± 0.11 & 0.78 ± 0.07 & 0.73 ± 0.11 & 0.73 ± 0.08 & \textbf{0.91 ± 0.09} \\
Labyrinth2 & 0.51 ± 0.07 & 0.73 ± 0.08 & 0.93 ± 0.06 & \textbf{0.95 ± 0.02} & 0.79 ± 0.12 \\
StandardMaze & 0.63 ± 0.09 & 0.84 ± 0.06 & 0.98 ± 0.02 & \textbf{1.00 ± 0.00} & \textbf{1.00 ± 0.00} \\
StandardMaze2 & 0.63 ± 0.10 & 0.88 ± 0.07 & 0.86 ± 0.08 & 0.99 ± 0.01 & \textbf{1.00 ± 0.00} \\
StandardMaze3 & 0.91 ± 0.05 & 0.99 ± 0.01 & \textbf{1.00 ± 0.00} & 0.98 ± 0.02 & \textbf{1.00 ± 0.00} \\
\midrule
Mean & 0.75 ± 0.03 & 0.87 ± 0.03 & 0.92 ± 0.02 & 0.91 ± 0.01 & \textbf{0.96 ± 0.03} \\
\bottomrule
\end{tabular}
\end{table}

\newpage

\subsubsection*{Key Minigrid 13x13 Results}
\begin{table}[H]
\centering
\caption{Key Minigrid 13x13 Solve Rate (1)}
\label{tab:Key_Minigrid_Solve_Rate_1}
\begin{tabular}{llllll}
\toprule
Level & Initial Gen - MNA & DR & SFL & PLR - MaxMC & ACCEL - MaxMC \\
\midrule
SixteenRooms\_Key & 0.18 ± 0.12 & 0.59 ± 0.12 & 0.58 ± 0.12 & 0.41 ± 0.11 & 0.83 ± 0.04 \\
SixteenRooms2\_Key & 0.12 ± 0.12 & 0.55 ± 0.13 & 0.83 ± 0.10 & 0.19 ± 0.05 & 0.83 ± 0.08 \\
FourRooms\_Key & 0.03 ± 0.02 & 0.55 ± 0.13 & \textbf{0.98 ± 0.02} & 0.10 ± 0.08 & 0.96 ± 0.02 \\
FourRooms2\_Key & 0.05 ± 0.04 & 0.88 ± 0.06 & 0.93 ± 0.07 & 0.79 ± 0.10 & 0.98 ± 0.02 \\
FourRooms3\_Key & 0.29 ± 0.14 & 0.51 ± 0.14 & 0.85 ± 0.06 & 0.10 ± 0.06 & 0.95 ± 0.05 \\
StandardMaze\_Key & 0.01 ± 0.01 & 0.09 ± 0.04 & 0.35 ± 0.12 & 0.24 ± 0.07 & 0.20 ± 0.10 \\
StandardMaze2\_Key & 0.00 ± 0.00 & 0.61 ± 0.13 & 0.60 ± 0.14 & 0.40 ± 0.11 & 0.26 ± 0.07 \\
StandardMaze3\_Key & 0.00 ± 0.00 & 0.36 ± 0.14 & 0.30 ± 0.14 & 0.09 ± 0.06 & 0.75 ± 0.05 \\
\midrule
Mean & 0.08 ± 0.04 & 0.52 ± 0.05 & 0.68 ± 0.04 & 0.29 ± 0.03 & 0.72 ± 0.02 \\
\bottomrule
\end{tabular}
\end{table}

\begin{table}[H]
\centering
\caption{Key Minigrid 13x13 Solve Rate (2)}
\label{tab:Key_Minigrid_Solve_Rate_2}
\begin{tabular}{llllll}
\toprule
Level & PLR - PVL & ACCEL - PVL & PLR - MNA & ACCEL - MNA & DEGen - MNA \\
\midrule
SixteenRooms\_Key & 0.15 ± 0.10 & 0.55 ± 0.04 & 0.95 ± 0.03 & 0.99 ± 0.01 & \textbf{1.00 ± 0.00} \\
SixteenRooms2\_Key & 0.00 ± 0.00 & 0.63 ± 0.10 & 0.90 ± 0.06 & 0.88 ± 0.10 & \textbf{1.00 ± 0.00} \\
FourRooms\_Key & 0.00 ± 0.00 & 0.60 ± 0.12 & 0.95 ± 0.04 & \textbf{0.98 ± 0.02} & 0.94 ± 0.05 \\
FourRooms2\_Key & 0.06 ± 0.03 & 0.80 ± 0.06 & \textbf{1.00 ± 0.00} & 0.99 ± 0.01 & \textbf{1.00 ± 0.00} \\
FourRooms3\_Key & 0.00 ± 0.00 & 0.66 ± 0.09 & 0.95 ± 0.03 & 0.98 ± 0.02 & \textbf{1.00 ± 0.00} \\
StandardMaze\_Key & 0.00 ± 0.00 & 0.00 ± 0.00 & 0.46 ± 0.10 & 0.85 ± 0.08 & \textbf{0.93 ± 0.06} \\
StandardMaze2\_Key & 0.00 ± 0.00 & 0.06 ± 0.02 & 0.73 ± 0.11 & \textbf{0.75 ± 0.08} & 0.63 ± 0.14 \\
StandardMaze3\_Key & 0.00 ± 0.00 & 0.30 ± 0.09 & 0.78 ± 0.07 & 0.84 ± 0.09 & \textbf{0.94 ± 0.05} \\
\midrule
Mean & 0.03 ± 0.01 & 0.45 ± 0.01 & 0.84 ± 0.01 & 0.90 ± 0.03 & \textbf{0.93 ± 0.02} \\
\bottomrule
\end{tabular}
\end{table}

\subsubsection*{Key Minigrid 17x17 Results}

\begin{table}[H]
\centering
\caption{Key Minigrid 17x17 Solve Rate}
\label{tab:Key_Minigrid_17_Solve_Rate}
\begin{tabular}{llll}
\toprule
Level & PLR - MNA & ACCEL - MNA & DEGen - MNA \\
\midrule
SixteenRooms\_Key & 0.99 ± 0.01 & 0.86 ± 0.05 & \textbf{1.00 ± 0.00} \\
SixteenRooms2\_Key & 0.86 ± 0.03 & 0.40 ± 0.13 & \textbf{0.95 ± 0.05} \\
FourRooms\_Key & 0.80 ± 0.11 & 0.64 ± 0.09 & \textbf{0.94 ± 0.05} \\
FourRooms2\_Key & 0.92 ± 0.05 & 0.81 ± 0.10 & \textbf{0.95 ± 0.04} \\
FourRooms3\_Key & 0.88 ± 0.07 & 0.65 ± 0.09 & \textbf{0.94 ± 0.04} \\
StandardMaze\_Key & \textbf{0.69 ± 0.13} & 0.19 ± 0.08 & 0.23 ± 0.05 \\
StandardMaze2\_Key & 0.29 ± 0.09 & 0.21 ± 0.07 & \textbf{0.50 ± 0.07} \\
StandardMaze3\_Key & 0.73 ± 0.09 & \textbf{0.83 ± 0.06} & 0.73 ± 0.09 \\
\midrule
Mean & 0.77 ± 0.04 & 0.57 ± 0.03 & \textbf{0.78 ± 0.03} \\
\bottomrule
\end{tabular}
\end{table}

\newpage

\subsubsection*{Key Minigrid 21x21 Results}

\begin{table}[H]
\centering
\caption{Key Minigrid 21x21 Solve Rate}
\label{tab:Key_Minigrid_21_Solve_Rate}
\begin{tabular}{llll}
\toprule
Level & PLR - MNA & ACCEL - MNA & DEGen - MNA \\
\midrule
SixteenRooms\_Key & 0.60 ± 0.11 & 0.60 ± 0.12 & \textbf{1.00 ± 0.00} \\
SixteenRooms2\_Key & 0.40 ± 0.09 & 0.45 ± 0.11 & \textbf{0.99 ± 0.01} \\
FourRooms\_Key & 0.21 ± 0.10 & 0.41 ± 0.11 & \textbf{0.99 ± 0.01} \\
FourRooms2\_Key & \textbf{0.93 ± 0.04} & 0.59 ± 0.09 & \textbf{0.93 ± 0.05} \\
FourRooms3\_Key & 0.35 ± 0.12 & 0.19 ± 0.08 & \textbf{0.98 ± 0.02} \\
StandardMaze\_Key & 0.34 ± 0.11 & 0.06 ± 0.03 & \textbf{0.56 ± 0.11} \\
StandardMaze2\_Key & 0.26 ± 0.12 & 0.00 ± 0.00 & \textbf{0.39 ± 0.12} \\
StandardMaze3\_Key & 0.38 ± 0.11 & 0.16 ± 0.07 & \textbf{0.61 ± 0.11} \\
\midrule
Mean & 0.43 ± 0.05 & 0.31 ± 0.04 & \textbf{0.80 ± 0.03} \\
\bottomrule
\end{tabular}
\end{table}

\newpage

\section{Sokoban Environment}

We have additionally performed experiments in a Sokoban-style environment. As in standard Sokoban, the agent aims to get all boxes to their storage locations. The agent receives a sparse reward when completing a level inversly proportional to how many steps were required to complete the level. For this domain, similarly to minigrid the agent has 5x5 observation space ahead of the agent, and an action space of move \textit{forward}, turn \textit{left} or turn \textit{right}. The agent also has a \textit{reset} action that resets the agent and boxes to their starting locations.

We used 9x9 levels for training - for DR, PLR and SFL, the random level generator generated levels that had 15 walls, and between 1 and 10 boxes. For DEGen, the generator could fill each newly observed cell in the environment with an empty square/wall/box/storage location/box on storage location. For all methods, we used identical hyperparameters to the minigrid environment (See Table \ref{tab:Learning_Hyperparameters}).

\subsection{Sokoban Results}

\begin{figure}[ht]
  \centering
  \includegraphics[width=0.8\textwidth]{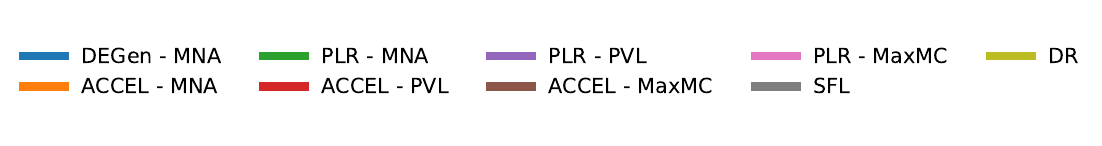}\\[-1.5em]
  
  \subfigure[Mean Return]{
    \includegraphics[width=0.48\textwidth]{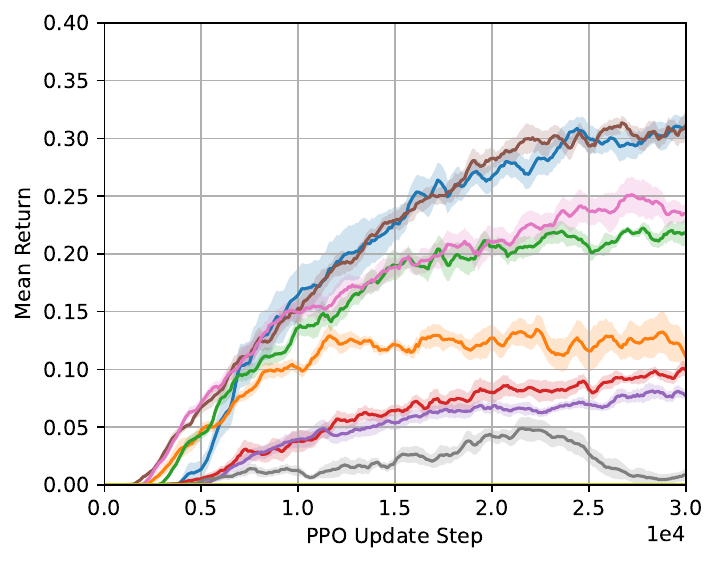}
    \label{fig:Sokoban_Mean_Return}
  }
  \subfigure[Mean Solve Rate]{
    \includegraphics[width=0.48\textwidth]{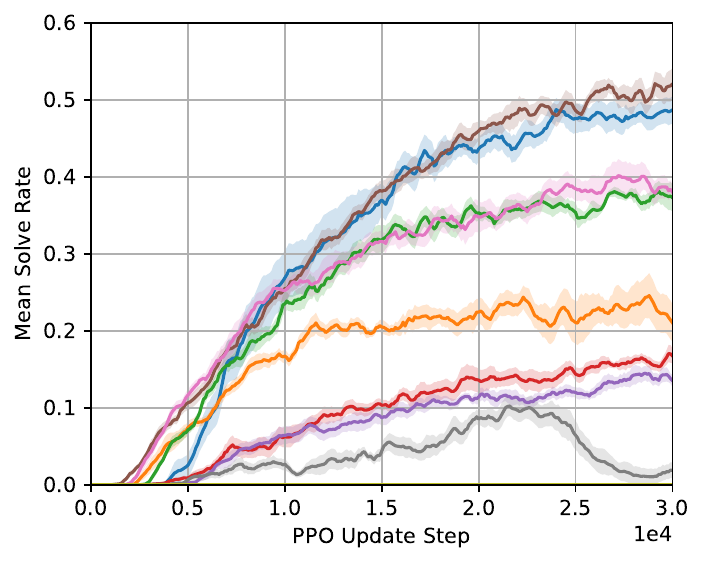}
    \label{fig:Sokoban_Mean_Solve_Rate}
  }
  \caption{Sokoban zero-shot performance on hand-designed test set, showing mean and standard error across 8 runs.}
  \label{fig:Sokoban_Results}
\end{figure}

From the results shown in Figure \ref{fig:Sokoban_Results} and Tables \ref{tab:Sokoban_Solve_Rate_1} and \ref{tab:Sokoban_Solve_Rate_2}, we see a large range in the performance of the various UED methods. We see that ACCEL using MaxMC is the best performing method, although comparable performance is achieved using DEGen. In Sokoban, we see that MaxMC outperforms MNA in the replay-based methods, with a substantial performance difference when used with ACCEL. Sokoban presents unique challenges compared to the previous environments used in this work. Primarily, the majority of randomly-sampled levels tend to be impossible, given that it only takes one box or storage location to be unreachable to make the entire level unsolvable. However, it is also likely that a high proportion of the solvable randomly-sampled levels will be difficult. 

As all impossible levels will necessarily score zero with both MaxMC and MNA, We hypothesise that MaxMC may be a better metric than MNA in domains where difficult levels represent a higher proportion of the non-zero scoring randomly-sampled levels. The high performance of ACCEL in Sokoban is likely due to the clear difficulty scaling that can be achieved with ACCEL-like level evolution. In our implementation of ACCEL, one of the possible mutations is to add or remove a box/storage-location pair. This enables gradual difficulty evolution in Sokoban, which is less likely in the minigrid enviroments. As such, ACCEL is highly effective at generating an curriculum.

We do however see in Tables \ref{tab:Sokoban_Solve_Rate_1} and \ref{tab:Sokoban_Solve_Rate_2} that a number of levels, those marked in \textit{italics}, are not solved by any method. This suggests that there is room for future work to enable zero-shot performance on more difficult levels, and that Sokoban may be an interesting environment for future UED research.

\newpage

\subsection{Sokoban Zero-shot Transfer Levels}

For the zero-shot transfer set, we have used the first 20 Sokoban Jr levels that do not exceed 13x13 in size. 

\begin{figure}[H]
\vskip 0.2in
\begin{center}
\centerline{\includegraphics[width=0.95\textwidth]{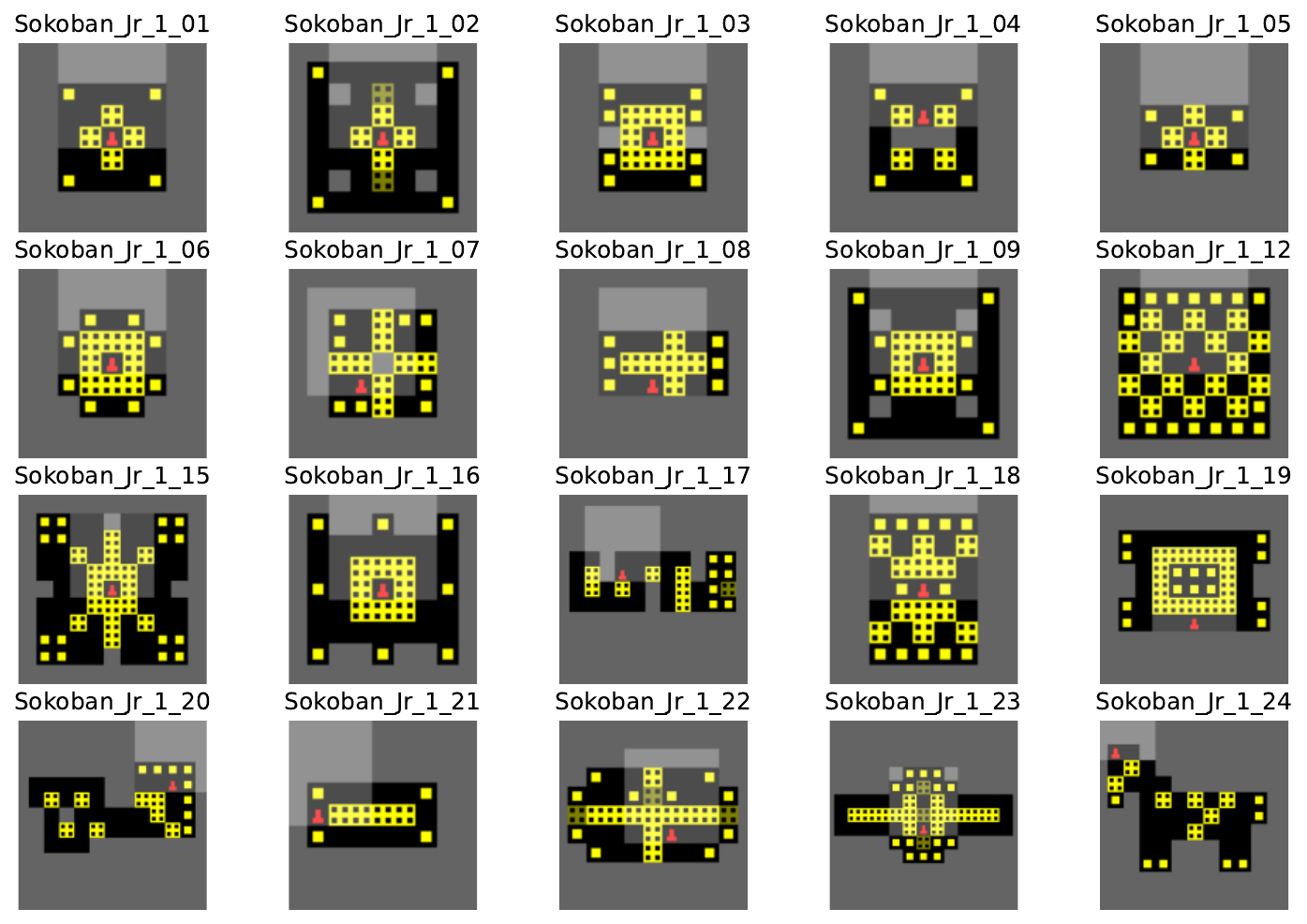}}
\caption{Hand designed evaluation levels for sokoban}
\label{fig:Sokoban_Eval_Levels}
\end{center}
\vskip -0.2in
\end{figure}

\newpage

\subsection{Sokoban Tabular Results}
\begin{table}[H]
\centering
\caption{Sokoban Solve Rate (1)}
\label{tab:Sokoban_Solve_Rate_1}
\begin{tabular}{lllll}
\toprule
Level & DR & SFL & PLR - MaxMC & ACCEL - MaxMC \\
\midrule
Sokoban\_Jr\_1\_01 & 0.00 ± 0.00 & 0.23 ± 0.15 & \textbf{1.00 ± 0.00} & \textbf{1.00 ± 0.00} \\
Sokoban\_Jr\_1\_02 & 0.00 ± 0.00 & 0.18 ± 0.12 & 0.94 ± 0.05 & 0.98 ± 0.02 \\
Sokoban\_Jr\_1\_03 & 0.00 ± 0.00 & 0.00 ± 0.00 & 0.46 ± 0.13 & \textbf{0.83 ± 0.12} \\
Sokoban\_Jr\_1\_04 & 0.00 ± 0.00 & 0.00 ± 0.00 & 0.16 ± 0.07 & 0.30 ± 0.09 \\
Sokoban\_Jr\_1\_05 & 0.00 ± 0.00 & 0.00 ± 0.00 & 0.42 ± 0.14 & 0.45 ± 0.12 \\
Sokoban\_Jr\_1\_06 & 0.00 ± 0.00 & 0.00 ± 0.00 & 0.74 ± 0.11 & 0.75 ± 0.11 \\
Sokoban\_Jr\_1\_07 & 0.00 ± 0.00 & 0.00 ± 0.00 & 0.60 ± 0.14 & 0.96 ± 0.03 \\
Sokoban\_Jr\_1\_08 & 0.00 ± 0.00 & 0.00 ± 0.00 & 0.21 ± 0.09 & \textbf{0.50 ± 0.13} \\
Sokoban\_Jr\_1\_09 & 0.00 ± 0.00 & 0.00 ± 0.00 & 0.65 ± 0.05 & 0.81 ± 0.04 \\
Sokoban\_Jr\_1\_12 & 0.00 ± 0.00 & 0.00 ± 0.00 & 0.00 ± 0.00 & 0.29 ± 0.06 \\
\textit{Sokoban\_Jr\_1\_15} & \textit{0.00 ± 0.00} & \textit{0.00 ± 0.00} & \textit{0.00 ± 0.00} & \textit{0.00 ± 0.00} \\
Sokoban\_Jr\_1\_16 & 0.00 ± 0.00 & 0.00 ± 0.00 & 0.93 ± 0.04 & \textbf{0.96 ± 0.02} \\
\textit{Sokoban\_Jr\_1\_17} & \textit{0.00 ± 0.00} & \textit{0.00 ± 0.00} & \textit{0.00 ± 0.00} & \textit{0.00 ± 0.00} \\
Sokoban\_Jr\_1\_18 & 0.00 ± 0.00 & 0.00 ± 0.00 & 0.04 ± 0.03 & 0.46 ± 0.10 \\
Sokoban\_Jr\_1\_19 & 0.00 ± 0.00 & 0.00 ± 0.00 & 0.00 ± 0.00 & \textbf{0.23 ± 0.11} \\
\textit{Sokoban\_Jr\_1\_20} & \textit{0.00 ± 0.00} & \textit{0.00 ± 0.00} & \textit{0.00 ± 0.00} & \textit{0.00 ± 0.00} \\
Sokoban\_Jr\_1\_21 & 0.00 ± 0.00 & 0.00 ± 0.00 & 0.96 ± 0.03 & \textbf{1.00 ± 0.00} \\
Sokoban\_Jr\_1\_22 & 0.00 ± 0.00 & 0.00 ± 0.00 & 0.14 ± 0.02 & \textbf{0.64 ± 0.10} \\
\textit{Sokoban\_Jr\_1\_23} & \textit{0.00 ± 0.00} & \textit{0.00 ± 0.00} & \textit{0.00 ± 0.00} & \textit{0.00 ± 0.00} \\
Sokoban\_Jr\_1\_24 & 0.00 ± 0.00 & 0.00 ± 0.00 & 0.04 ± 0.03 & \textbf{0.14 ± 0.05} \\
\midrule
Mean & 0.00 ± 0.00 & 0.02 ± 0.01 & 0.36 ± 0.01 & \textbf{0.51 ± 0.02} \\
\bottomrule
\end{tabular}
\end{table}

\begin{table}[H]
\centering
\caption{Sokoban Solve Rate (2)}
\label{tab:Sokoban_Solve_Rate_2}
\begin{tabular}{llllll}
\toprule
Level & PLR - PVL & ACCEL - PVL & PLR - MNA & ACCEL - MNA & DEGen - MNA \\
\midrule
Sokoban\_Jr\_1\_01 & \textbf{1.00 ± 0.00} & 0.99 ± 0.01 & 0.98 ± 0.02 & 0.89 ± 0.11 & \textbf{1.00 ± 0.00} \\
Sokoban\_Jr\_1\_02 & 0.24 ± 0.05 & 0.46 ± 0.06 & 0.98 ± 0.02 & 0.49 ± 0.07 & \textbf{1.00 ± 0.00} \\
Sokoban\_Jr\_1\_03 & 0.09 ± 0.07 & 0.25 ± 0.10 & 0.71 ± 0.11 & 0.15 ± 0.08 & 0.49 ± 0.14 \\
Sokoban\_Jr\_1\_04 & 0.00 ± 0.00 & 0.00 ± 0.00 & \textbf{0.44 ± 0.09} & 0.14 ± 0.08 & 0.35 ± 0.09 \\
Sokoban\_Jr\_1\_05 & 0.10 ± 0.10 & 0.01 ± 0.01 & \textbf{0.58 ± 0.12} & 0.48 ± 0.12 & 0.24 ± 0.16 \\
Sokoban\_Jr\_1\_06 & 0.05 ± 0.03 & 0.03 ± 0.02 & 0.50 ± 0.14 & 0.46 ± 0.12 & \textbf{0.94 ± 0.03} \\
Sokoban\_Jr\_1\_07 & 0.19 ± 0.13 & 0.51 ± 0.16 & 0.35 ± 0.09 & 0.19 ± 0.07 & \textbf{1.00 ± 0.00} \\
Sokoban\_Jr\_1\_08 & 0.00 ± 0.00 & 0.00 ± 0.00 & 0.15 ± 0.11 & 0.00 ± 0.00 & \textbf{0.50 ± 0.15} \\
Sokoban\_Jr\_1\_09 & 0.14 ± 0.04 & 0.13 ± 0.07 & 0.68 ± 0.06 & 0.18 ± 0.06 & \textbf{0.85 ± 0.06} \\
Sokoban\_Jr\_1\_12 & 0.00 ± 0.00 & 0.00 ± 0.00 & 0.00 ± 0.00 & 0.00 ± 0.00 & \textbf{0.30 ± 0.15} \\
\textit{Sokoban\_Jr\_1\_15} & \textit{0.00 ± 0.00} & \textit{0.00 ± 0.00} & \textit{0.00 ± 0.00} & \textit{0.00 ± 0.00} & \textit{0.00 ± 0.00} \\
Sokoban\_Jr\_1\_16 & 0.25 ± 0.05 & 0.13 ± 0.06 & 0.95 ± 0.03 & 0.60 ± 0.08 & 0.95 ± 0.03 \\
\textit{Sokoban\_Jr\_1\_17} & \textit{0.00 ± 0.00} & \textit{0.00 ± 0.00} & \textit{0.00 ± 0.00} & \textit{0.00 ± 0.00} & \textit{0.00 ± 0.00} \\
Sokoban\_Jr\_1\_18 & 0.00 ± 0.00 & 0.03 ± 0.02 & 0.00 ± 0.00 & 0.00 ± 0.00 & \textbf{0.69 ± 0.14} \\
Sokoban\_Jr\_1\_19 & 0.00 ± 0.00 & 0.00 ± 0.00 & 0.00 ± 0.00 & 0.00 ± 0.00 & 0.10 ± 0.06 \\
\textit{Sokoban\_Jr\_1\_20} & \textit{0.00 ± 0.00} & \textit{0.00 ± 0.00} & \textit{0.00 ± 0.00} & \textit{0.00 ± 0.00} & \textit{0.00 ± 0.00} \\
Sokoban\_Jr\_1\_21 & 0.51 ± 0.09 & 0.56 ± 0.12 & 0.76 ± 0.09 & 0.68 ± 0.12 & 0.99 ± 0.01 \\
Sokoban\_Jr\_1\_22 & 0.00 ± 0.00 & 0.06 ± 0.05 & 0.09 ± 0.05 & 0.00 ± 0.00 & 0.30 ± 0.11 \\
\textit{Sokoban\_Jr\_1\_23} & \textit{0.00 ± 0.00} & \textit{0.00 ± 0.00} & \textit{0.00 ± 0.00} & \textit{0.00 ± 0.00} & \textit{0.00 ± 0.00} \\
Sokoban\_Jr\_1\_24 & 0.00 ± 0.00 & 0.01 ± 0.01 & 0.05 ± 0.04 & 0.00 ± 0.00 & 0.09 ± 0.03 \\
\midrule
Mean & 0.13 ± 0.01 & 0.16 ± 0.01 & 0.36 ± 0.01 & 0.21 ± 0.02 & 0.49 ± 0.02 \\
\bottomrule
\end{tabular}
\end{table}

\newpage

\section{Additional Results} \label{app:Additional_Results}

\subsection{MNA and Existing Methods} \label{app:MNA_Existing_Methods}

To directly examine the effectiveness of MNA compared to existing regret metics, we show zero-shot performance of ACCEL and PLR using MNA, PVL and MaxMC. 

\subsubsection*{Minigrid}

\begin{figure}[ht]
  \centering
  \includegraphics[width=0.55\textwidth]{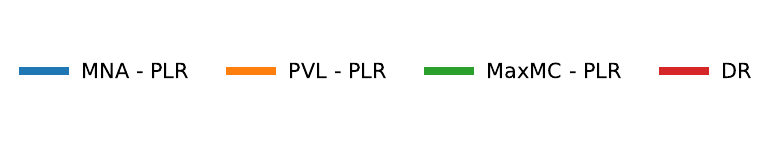}\\[-2em]
  
  \subfigure[Mean Return]{
    \includegraphics[width=0.48\textwidth]{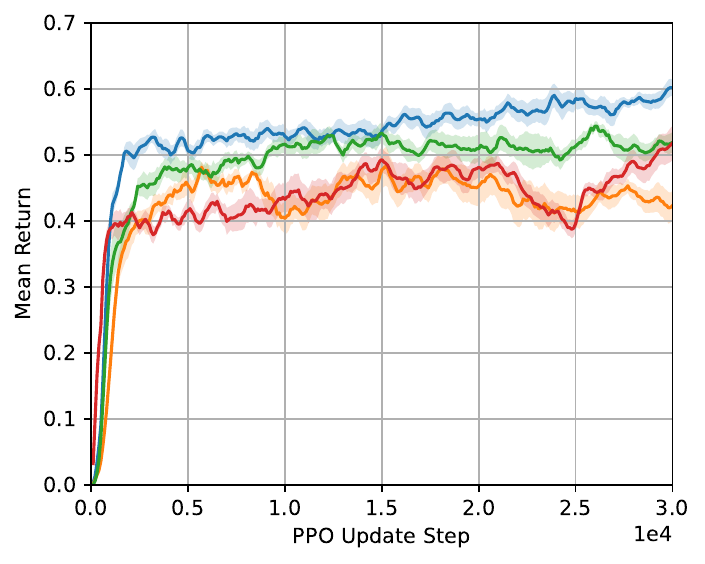}
  }
  \subfigure[Mean Solve Rate]{
    \includegraphics[width=0.48\textwidth]{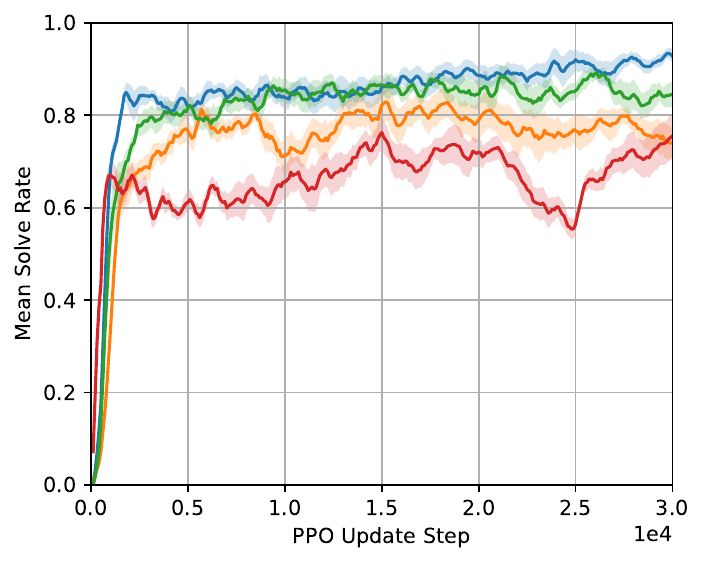}
  }
  \caption{Minigrid - comparison of PLR performance using different metrics}
  \label{fig:Minigrid_PLR}
\end{figure}

\begin{figure}[ht]
  \centering
  \includegraphics[width=0.55\textwidth]{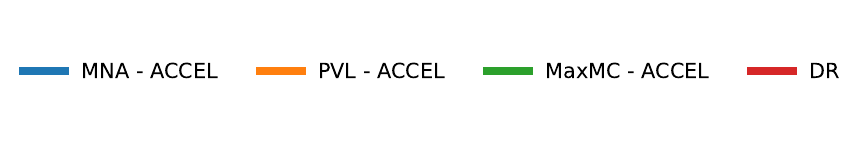}\\[-2em]
  
  \subfigure[Mean Return]{
    \includegraphics[width=0.48\textwidth]{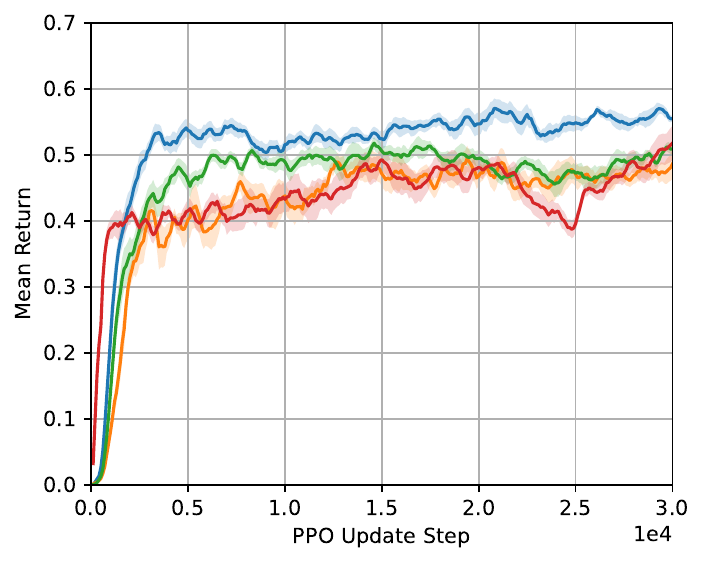}
  }
  \subfigure[Mean Solve Rate]{
    \includegraphics[width=0.48\textwidth]{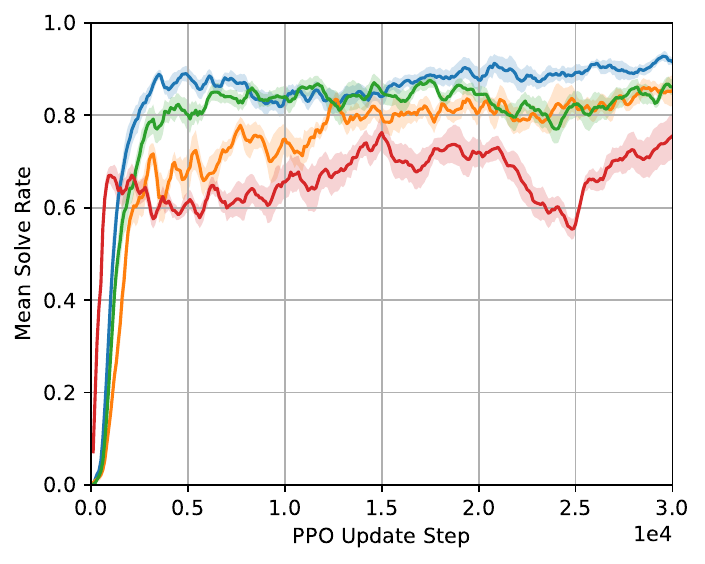}
  }
  \caption{Minigrid - comparison of ACCEL performance using different metrics}
  \label{fig:Minigrid_ACCEL}
\end{figure}

In Figures \ref{fig:Minigrid_PLR} and \ref{fig:Minigrid_ACCEL}, we illustrate the relative performance of each of these metrics in the standard minigrid domain and show that MNA clearly outperforms other metrics, whether using either PLR or ACCEL. 

\newpage

\subsubsection*{Key Minigrid}
\begin{figure}[ht]
  \centering
  \includegraphics[width=0.55\textwidth]{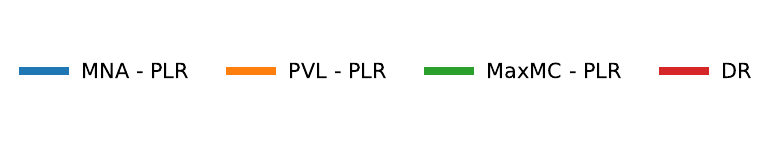}\\[-2em]
  
  \subfigure[Mean Return]{
    \includegraphics[width=0.48\textwidth]{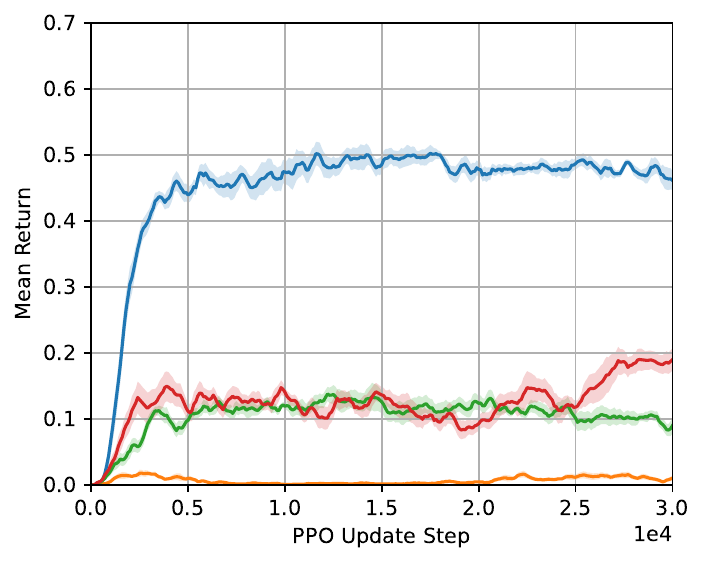}
  }
  \subfigure[Mean Solve Rate]{
    \includegraphics[width=0.48\textwidth]{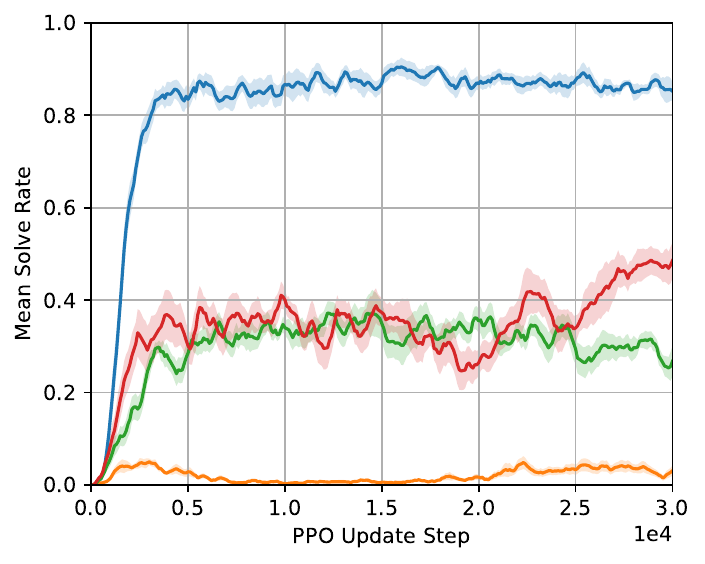}
  }
  \caption{Key Minigrid - comparison of PLR performance using different metrics}
  \label{fig:Key_Minigrid_PLR}
\end{figure}

\begin{figure}[ht]
  \centering
  \includegraphics[width=0.55\textwidth]{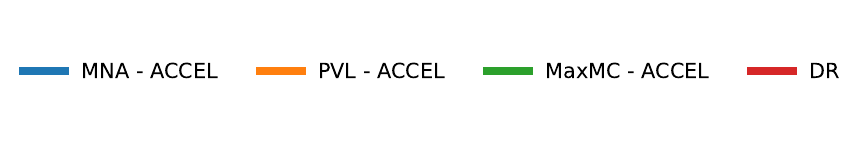}\\[-2em]
  
  \subfigure[Mean Return]{
    \includegraphics[width=0.48\textwidth]{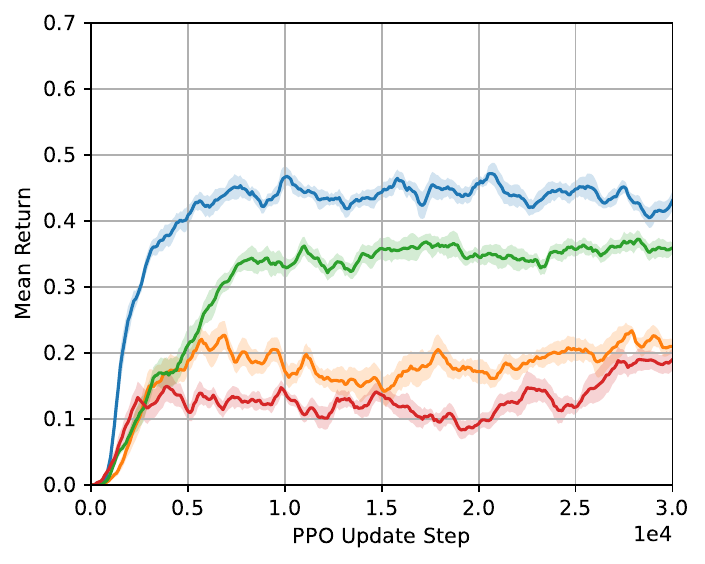}
  }
  \subfigure[Mean Solve Rate]{
    \includegraphics[width=0.48\textwidth]{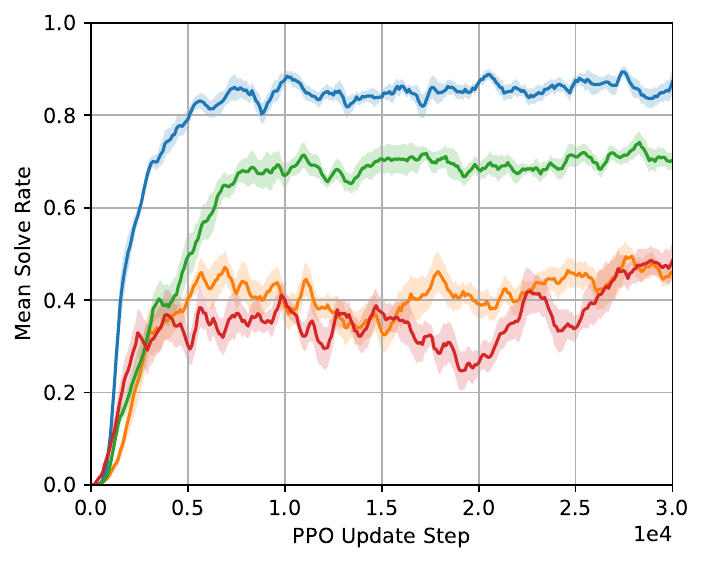}
  }
  \caption{Key Minigrid - comparison of ACCEL performance using different metrics}
  \label{fig:Key_Minigrid_ACCEL}
\end{figure}

In Figures \ref{fig:Key_Minigrid_PLR} and \ref{fig:Key_Minigrid_ACCEL}, we compare the same methods but on the key minigrid domain instead. Here, we again see that MNA outperforms existing methods - including a substantial performance improvement when using PLR. 

\newpage

\subsection{DEGen and Existing Regret Approximations} \label{app:DEGen_Regret_Approx_Compare}

\begin{figure}[ht]
  \centering
  \includegraphics[width=0.55\textwidth]{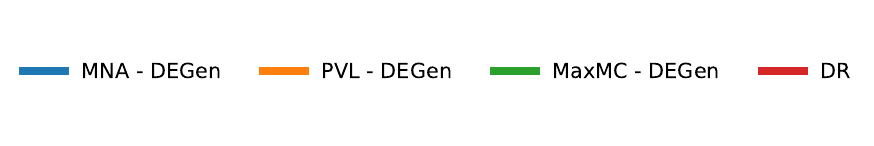}\\[-2em]
  
  \subfigure[Mean Return]{
    \includegraphics[width=0.48\textwidth]{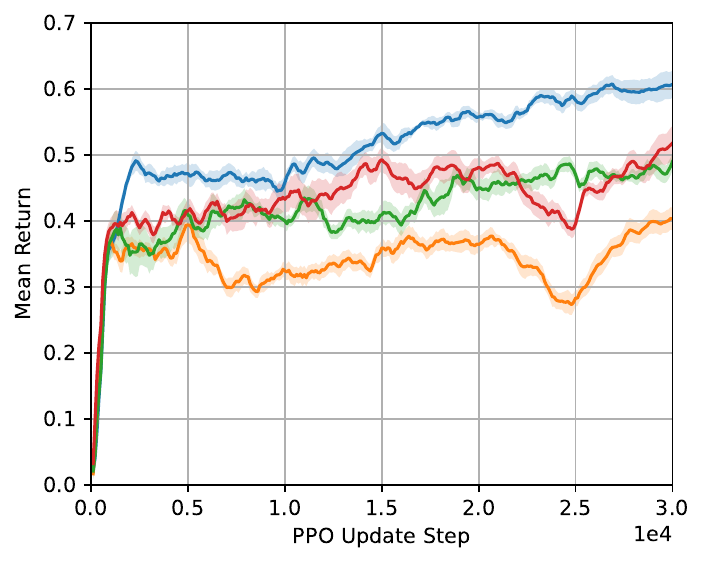}
  }
  \subfigure[Mean Solve Rate]{
    \includegraphics[width=0.48\textwidth]{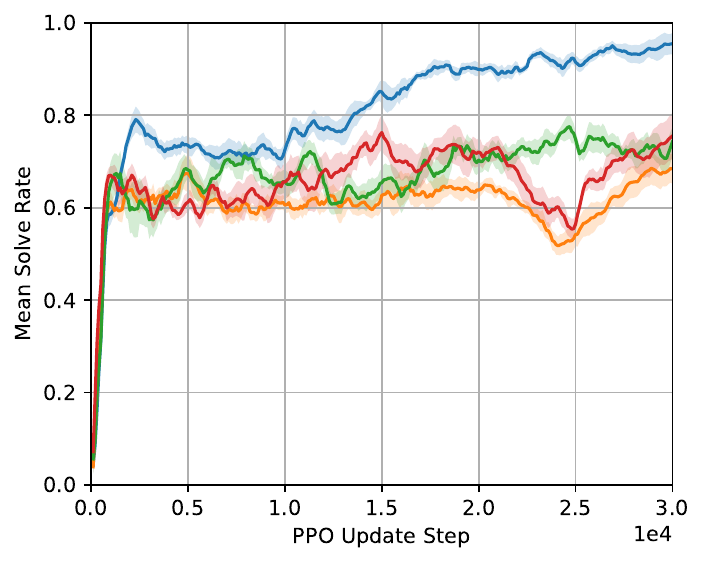}
  }
  \caption{Minigrid - comparison of DEGen performance using different metrics}
  \label{fig:Minigrid_DEGen}
\end{figure}

\begin{figure}[ht]
  \centering
  \includegraphics[width=0.55\textwidth]{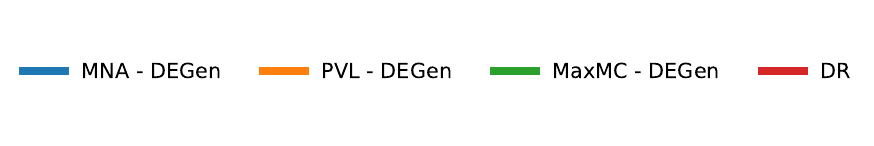}\\[-2em]
  
  \subfigure[Mean Return]{
    \includegraphics[width=0.48\textwidth]{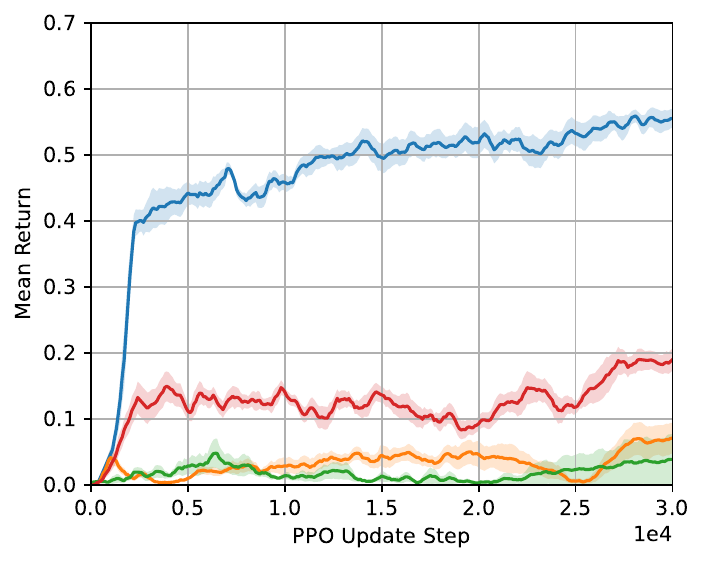}
  }
  \subfigure[Mean Solve Rate]{
    \includegraphics[width=0.48\textwidth]{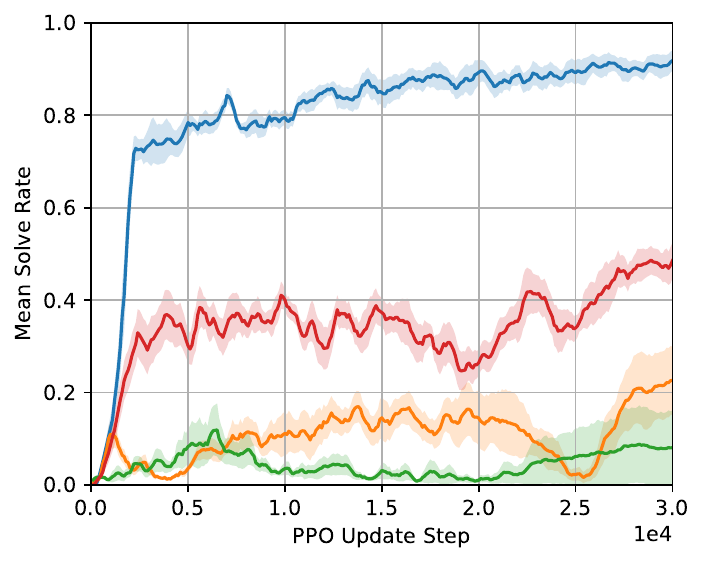}
  }
  \caption{Key Minigrid - comparison of DEGen performance using different metrics}
  \label{fig:Key_Minigrid_DEGen}
\end{figure}

In order to illustrate the ineffectiveness of existing regret approximations when used as optimisation objectives for training a teacher, we show the relative performance of DEGen using MNA, PVL and MaxMC. Figures \ref{fig:Minigrid_DEGen} and \ref{fig:Key_Minigrid_DEGen} show that MNA consistently outperforms PVL and MaxMC. We also see here that using a teacher trained using PVL and MaxMC results in at best equivalent, but generally worse, performance compared to naive domain randomisation. 

\newpage

\subsection{DEGen vs Initial Gen}

\begin{figure}[ht]
  \centering
  \includegraphics[width=0.55\textwidth]{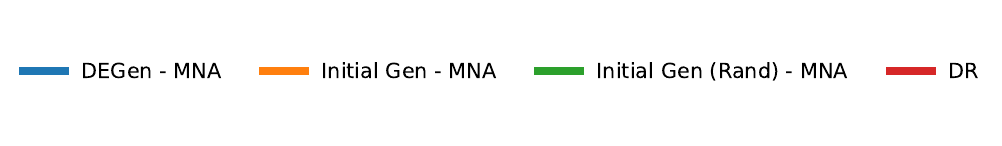}\\[-2em]
  
  \subfigure[Mean Return]{
    \includegraphics[width=0.48\textwidth]{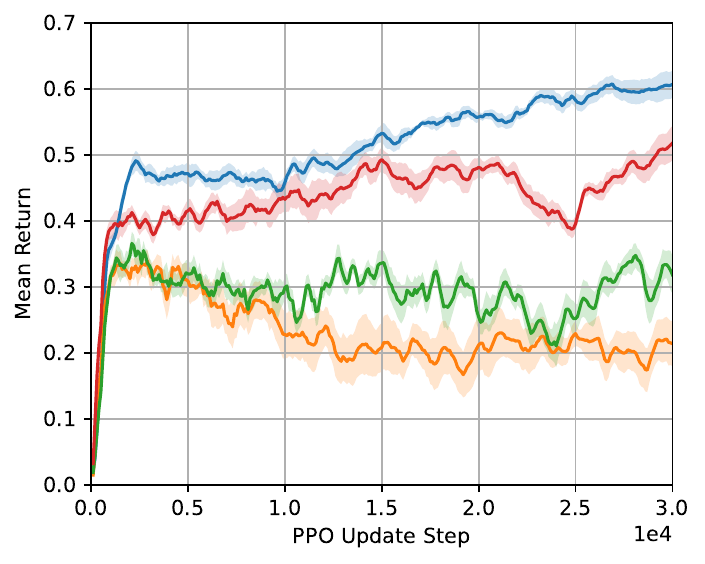}
  }
  \subfigure[Mean Solve Rate]{
    \includegraphics[width=0.48\textwidth]{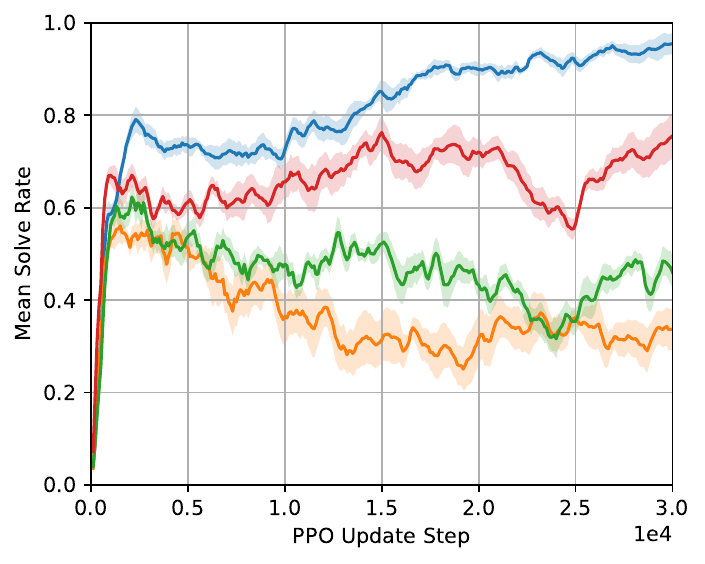}
  }
  \caption{Minigrid - comparison of Initial Gen and DEGen}
  \label{fig:Minigrid_Init_Gen}
\end{figure}

\begin{figure}[ht]
  \centering
  \includegraphics[width=0.55\textwidth]{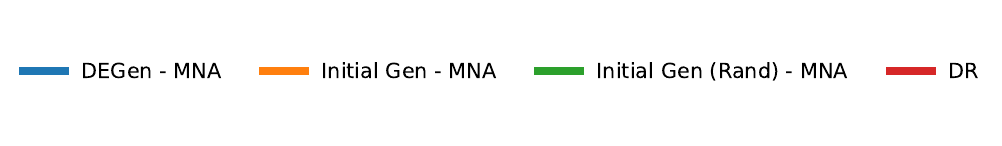}\\[-2em]
  
  \subfigure[Mean Return]{
    \includegraphics[width=0.48\textwidth]{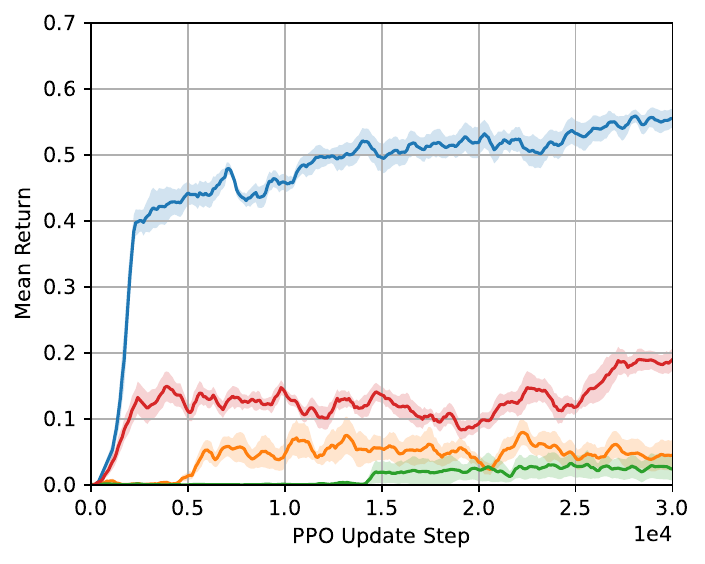}
  }
  \subfigure[Mean Solve Rate]{
    \includegraphics[width=0.48\textwidth]{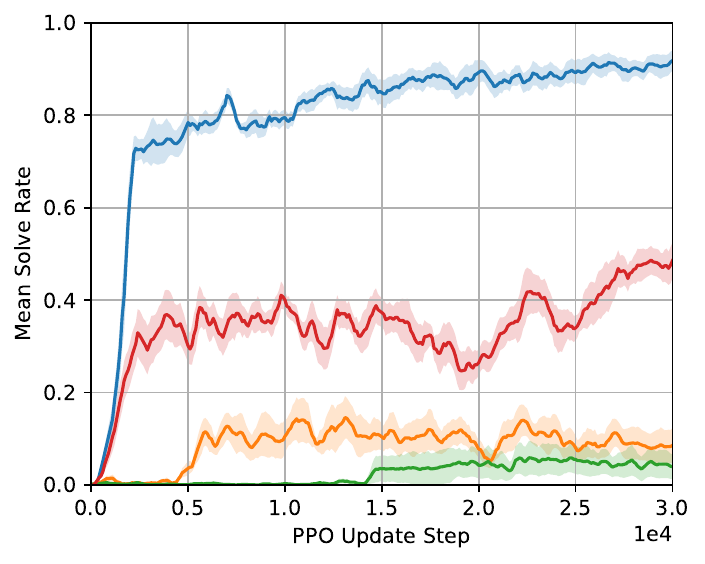}
  }
  \caption{Key Minigrid - comparison of Initial Gen and DEGen}
  \label{fig:Key_Minigrid_Init_Gen}
\end{figure}

In Figures \ref{fig:Minigrid_Init_Gen} and \ref{fig:Key_Minigrid_Init_Gen}, we show the performance of DEGen compared to the performance of a generator that generates the full level prior to student rollouts. We include both a standard level generator \textit{Initial Gen}, identical to the PAIRED generator in JaxUED \cite{coward2024jaxued}, as well as \textit{Initial Gen (Rand)}, which randomly places the agent in the level before the rest of the level is constructed. We see that both methods performance worse than domain randomisation. 

\newpage

\subsection{PAIRED}

Finally, we examine the performance of DEGen compared to PAIRED \cite{dennis2020emergent}. In standard minigrid, we see that PAIRED performs very similarly to domain randomisation, and worse than DEGen. Additionally, we see in the key minigrid domain the limitations of the PAIRED regret approximation. As PAIRED relies on the antagonist's performance to approximate the best possible level return, high scoring levels require the antagonist to perform well. However, if a level is challenging due to some obstacle the student has not previously encountered, it is likely that the antagonist will also perform poorly, given it has been trained on the same set of levels as the student. Therefore, the PAIRED generator is unlikely to generate levels requiring the antagonist to use the key, and so as the student agent has not encountered levels similar to the zero-shot hand-designed levels that require the key, zero-shot performance is extremely poor.

\begin{figure}[ht]
  \centering
  \includegraphics[width=0.5\textwidth]{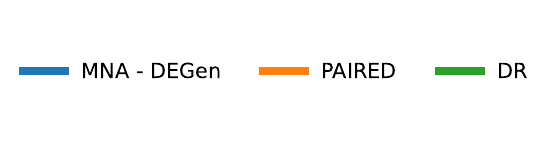}\\[-2em]
  
  \subfigure[Mean Return]{
    \includegraphics[width=0.48\textwidth]{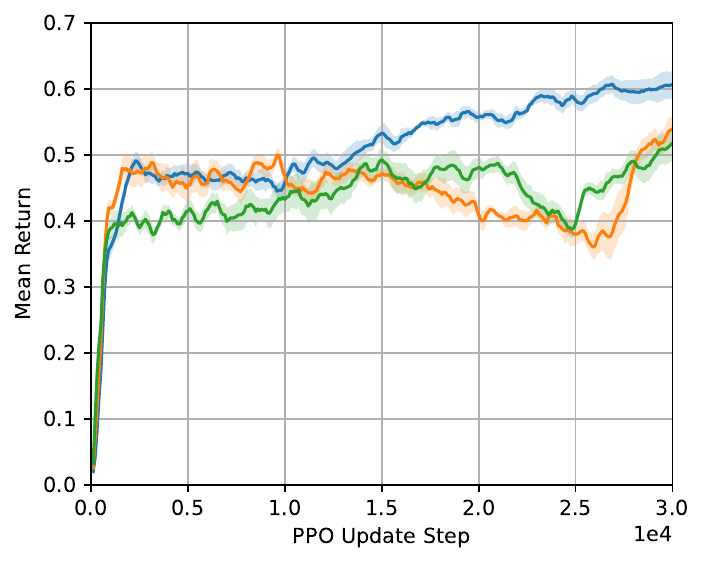}
  }
  \subfigure[Mean Solve Rate]{
    \includegraphics[width=0.48\textwidth]{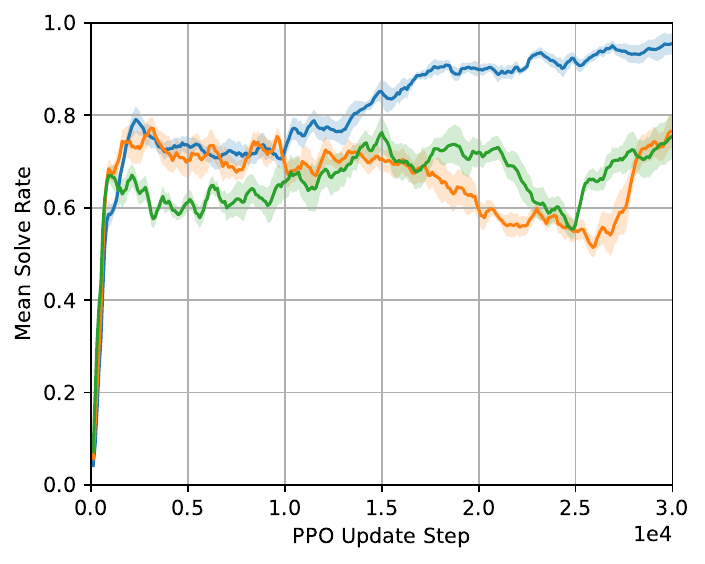}
  }
  \caption{Minigrid - comparison of PAIRED and DEGen}
  \label{fig:Minigrid_PAIRED}
\end{figure}

\begin{figure}[ht]
  \centering
  \includegraphics[width=0.5\textwidth]{graphics/MINIGRID_PAIRED_LEGEND.pdf}\\[-2em]
  
  \subfigure[Mean Return]{
    \includegraphics[width=0.48\textwidth]{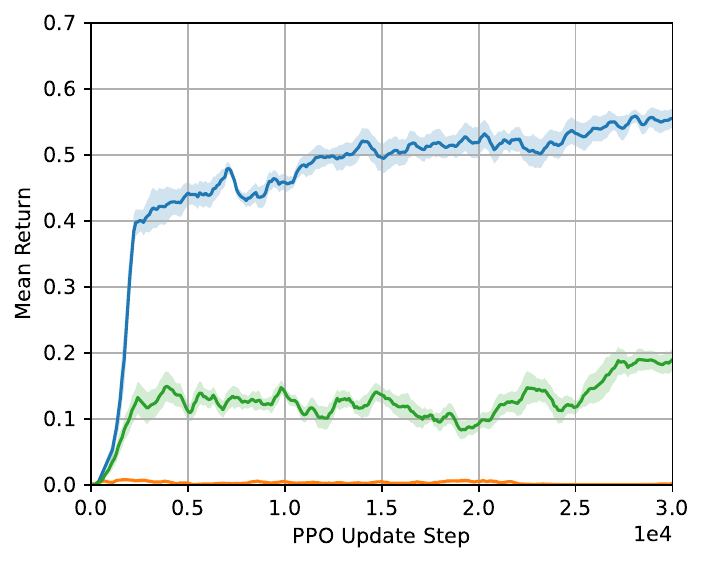}
  }
  \subfigure[Mean Solve Rate]{
    \includegraphics[width=0.48\textwidth]{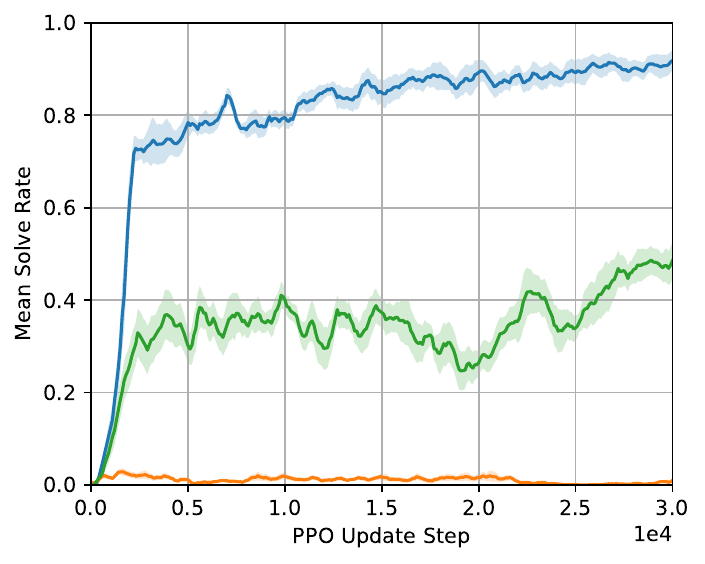}
  }
  \caption{Key Minigrid - comparison of PAIRED and DEGen}
  \label{fig:Key_Minigrid_PAIRED}
\end{figure}

\newpage

\subsection{Training Level Examples} \label{app:Training_Level_Examples}

We have included examples of levels generated by each method in the repository at \href{https://github.com/HarryMJMead/Dynamic-Environment-Generation-for-UED}{\texttt{https://github.com/HarryMJMead/Dynamic-Environment-Generation-for-UED}}. These levels were all sampled from the final training step.







\newpage

\section*{NeurIPS Paper Checklist}

\begin{enumerate}

\item {\bf Claims}
    \item[] Question: Do the main claims made in the abstract and introduction accurately reflect the paper's contributions and scope?
    \item[] Answer: \answerYes{} 
    \item[] Justification: The claims made in the abstract are accurately reflected in the description of our method and our results.
    \item[] Guidelines:
    \begin{itemize}
        \item The answer NA means that the abstract and introduction do not include the claims made in the paper.
        \item The abstract and/or introduction should clearly state the claims made, including the contributions made in the paper and important assumptions and limitations. A No or NA answer to this question will not be perceived well by the reviewers. 
        \item The claims made should match theoretical and experimental results, and reflect how much the results can be expected to generalize to other settings. 
        \item It is fine to include aspirational goals as motivation as long as it is clear that these goals are not attained by the paper. 
    \end{itemize}

\item {\bf Limitations}
    \item[] Question: Does the paper discuss the limitations of the work performed by the authors?
    \item[] Answer: \answerYes{}{} 
    \item[] Justification: We have discussed the limitations of this work in Section \ref{sec:Limitations}.
    \item[] Guidelines:
    \begin{itemize}
        \item The answer NA means that the paper has no limitation while the answer No means that the paper has limitations, but those are not discussed in the paper. 
        \item The authors are encouraged to create a separate "Limitations" section in their paper.
        \item The paper should point out any strong assumptions and how robust the results are to violations of these assumptions (e.g., independence assumptions, noiseless settings, model well-specification, asymptotic approximations only holding locally). The authors should reflect on how these assumptions might be violated in practice and what the implications would be.
        \item The authors should reflect on the scope of the claims made, e.g., if the approach was only tested on a few datasets or with a few runs. In general, empirical results often depend on implicit assumptions, which should be articulated.
        \item The authors should reflect on the factors that influence the performance of the approach. For example, a facial recognition algorithm may perform poorly when image resolution is low or images are taken in low lighting. Or a speech-to-text system might not be used reliably to provide closed captions for online lectures because it fails to handle technical jargon.
        \item The authors should discuss the computational efficiency of the proposed algorithms and how they scale with dataset size.
        \item If applicable, the authors should discuss possible limitations of their approach to address problems of privacy and fairness.
        \item While the authors might fear that complete honesty about limitations might be used by reviewers as grounds for rejection, a worse outcome might be that reviewers discover limitations that aren't acknowledged in the paper. The authors should use their best judgment and recognize that individual actions in favor of transparency play an important role in developing norms that preserve the integrity of the community. Reviewers will be specifically instructed to not penalize honesty concerning limitations.
    \end{itemize}

\item {\bf Theory assumptions and proofs}
    \item[] Question: For each theoretical result, does the paper provide the full set of assumptions and a complete (and correct) proof?
    \item[] Answer: \answerNA{} 
    \item[] Justification: This paper contains no theoretical results.
    \item[] Guidelines:
    \begin{itemize}
        \item The answer NA means that the paper does not include theoretical results. 
        \item All the theorems, formulas, and proofs in the paper should be numbered and cross-referenced.
        \item All assumptions should be clearly stated or referenced in the statement of any theorems.
        \item The proofs can either appear in the main paper or the supplemental material, but if they appear in the supplemental material, the authors are encouraged to provide a short proof sketch to provide intuition. 
        \item Inversely, any informal proof provided in the core of the paper should be complemented by formal proofs provided in appendix or supplemental material.
        \item Theorems and Lemmas that the proof relies upon should be properly referenced. 
    \end{itemize}

    \item {\bf Experimental result reproducibility}
    \item[] Question: Does the paper fully disclose all the information needed to reproduce the main experimental results of the paper to the extent that it affects the main claims and/or conclusions of the paper (regardless of whether the code and data are provided or not)?
    \item[] Answer: \answerYes{} 
    \item[] Justification: The full method used is outlined in Sections \ref{sec:DEGen} and \ref{sec:MNA}, and further detail is provided in Appendix \ref{app:Experimental_Setup}
    \item[] Guidelines:
    \begin{itemize}
        \item The answer NA means that the paper does not include experiments.
        \item If the paper includes experiments, a No answer to this question will not be perceived well by the reviewers: Making the paper reproducible is important, regardless of whether the code and data are provided or not.
        \item If the contribution is a dataset and/or model, the authors should describe the steps taken to make their results reproducible or verifiable. 
        \item Depending on the contribution, reproducibility can be accomplished in various ways. For example, if the contribution is a novel architecture, describing the architecture fully might suffice, or if the contribution is a specific model and empirical evaluation, it may be necessary to either make it possible for others to replicate the model with the same dataset, or provide access to the model. In general. releasing code and data is often one good way to accomplish this, but reproducibility can also be provided via detailed instructions for how to replicate the results, access to a hosted model (e.g., in the case of a large language model), releasing of a model checkpoint, or other means that are appropriate to the research performed.
        \item While NeurIPS does not require releasing code, the conference does require all submissions to provide some reasonable avenue for reproducibility, which may depend on the nature of the contribution. For example
        \begin{enumerate}
            \item If the contribution is primarily a new algorithm, the paper should make it clear how to reproduce that algorithm.
            \item If the contribution is primarily a new model architecture, the paper should describe the architecture clearly and fully.
            \item If the contribution is a new model (e.g., a large language model), then there should either be a way to access this model for reproducing the results or a way to reproduce the model (e.g., with an open-source dataset or instructions for how to construct the dataset).
            \item We recognize that reproducibility may be tricky in some cases, in which case authors are welcome to describe the particular way they provide for reproducibility. In the case of closed-source models, it may be that access to the model is limited in some way (e.g., to registered users), but it should be possible for other researchers to have some path to reproducing or verifying the results.
        \end{enumerate}
    \end{itemize}

\item {\bf Open access to data and code}
    \item[] Question: Does the paper provide open access to the data and code, with sufficient instructions to faithfully reproduce the main experimental results, as described in supplemental material?
    \item[] Answer: \answerYes{} 
    \item[] Justification: We include the code and instructions in the supplementary material
    \item[] Guidelines:
    \begin{itemize}
        \item The answer NA means that paper does not include experiments requiring code.
        \item Please see the NeurIPS code and data submission guidelines (\url{https://nips.cc/public/guides/CodeSubmissionPolicy}) for more details.
        \item While we encourage the release of code and data, we understand that this might not be possible, so “No” is an acceptable answer. Papers cannot be rejected simply for not including code, unless this is central to the contribution (e.g., for a new open-source benchmark).
        \item The instructions should contain the exact command and environment needed to run to reproduce the results. See the NeurIPS code and data submission guidelines (\url{https://nips.cc/public/guides/CodeSubmissionPolicy}) for more details.
        \item The authors should provide instructions on data access and preparation, including how to access the raw data, preprocessed data, intermediate data, and generated data, etc.
        \item The authors should provide scripts to reproduce all experimental results for the new proposed method and baselines. If only a subset of experiments are reproducible, they should state which ones are omitted from the script and why.
        \item At submission time, to preserve anonymity, the authors should release anonymized versions (if applicable).
        \item Providing as much information as possible in supplemental material (appended to the paper) is recommended, but including URLs to data and code is permitted.
    \end{itemize}

\item {\bf Experimental setting/details}
    \item[] Question: Does the paper specify all the training and test details (e.g., data splits, hyperparameters, how they were chosen, type of optimizer, etc.) necessary to understand the results?
    \item[] Answer: \answerYes{} 
    \item[] Justification: All addtional hyperparameters and experiment details are included in Appendix \ref{app:Hyperparameters}
    \item[] Guidelines:
    \begin{itemize}
        \item The answer NA means that the paper does not include experiments.
        \item The experimental setting should be presented in the core of the paper to a level of detail that is necessary to appreciate the results and make sense of them.
        \item The full details can be provided either with the code, in appendix, or as supplemental material.
    \end{itemize}

\item {\bf Experiment statistical significance}
    \item[] Question: Does the paper report error bars suitably and correctly defined or other appropriate information about the statistical significance of the experiments?
    \item[] Answer: \answerYes{} 
    \item[] Justification: All results are taken over multiple seeds with standard error
    \item[] Guidelines:
    \begin{itemize}
        \item The answer NA means that the paper does not include experiments.
        \item The authors should answer "Yes" if the results are accompanied by error bars, confidence intervals, or statistical significance tests, at least for the experiments that support the main claims of the paper.
        \item The factors of variability that the error bars are capturing should be clearly stated (for example, train/test split, initialization, random drawing of some parameter, or overall run with given experimental conditions).
        \item The method for calculating the error bars should be explained (closed form formula, call to a library function, bootstrap, etc.)
        \item The assumptions made should be given (e.g., Normally distributed errors).
        \item It should be clear whether the error bar is the standard deviation or the standard error of the mean.
        \item It is OK to report 1-sigma error bars, but one should state it. The authors should preferably report a 2-sigma error bar than state that they have a 96\% CI, if the hypothesis of Normality of errors is not verified.
        \item For asymmetric distributions, the authors should be careful not to show in tables or figures symmetric error bars that would yield results that are out of range (e.g. negative error rates).
        \item If error bars are reported in tables or plots, The authors should explain in the text how they were calculated and reference the corresponding figures or tables in the text.
    \end{itemize}

\item {\bf Experiments compute resources}
    \item[] Question: For each experiment, does the paper provide sufficient information on the computer resources (type of compute workers, memory, time of execution) needed to reproduce the experiments?
    \item[] Answer: \answerYes{} 
    \item[] Justification: Details on compute and time are provided in  Appendix \ref{app:Experimental_Setup}
    \item[] Guidelines:
    \begin{itemize}
        \item The answer NA means that the paper does not include experiments.
        \item The paper should indicate the type of compute workers CPU or GPU, internal cluster, or cloud provider, including relevant memory and storage.
        \item The paper should provide the amount of compute required for each of the individual experimental runs as well as estimate the total compute. 
        \item The paper should disclose whether the full research project required more compute than the experiments reported in the paper (e.g., preliminary or failed experiments that didn't make it into the paper). 
    \end{itemize}
    
\item {\bf Code of ethics}
    \item[] Question: Does the research conducted in the paper conform, in every respect, with the NeurIPS Code of Ethics \url{https://neurips.cc/public/EthicsGuidelines}?
    \item[] Answer: \answerYes{} 
    \item[] Justification: Our research conforms with the NeurIPS Code of Ethics.
    \item[] Guidelines:
    \begin{itemize}
        \item The answer NA means that the authors have not reviewed the NeurIPS Code of Ethics.
        \item If the authors answer No, they should explain the special circumstances that require a deviation from the Code of Ethics.
        \item The authors should make sure to preserve anonymity (e.g., if there is a special consideration due to laws or regulations in their jurisdiction).
    \end{itemize}

\item {\bf Broader impacts}
    \item[] Question: Does the paper discuss both potential positive societal impacts and negative societal impacts of the work performed?
    \item[] Answer: \answerNA{} 
    \item[] Justification: Our work is not expected to have direct societal impacts.
    \item[] Guidelines:
    \begin{itemize}
        \item The answer NA means that there is no societal impact of the work performed.
        \item If the authors answer NA or No, they should explain why their work has no societal impact or why the paper does not address societal impact.
        \item Examples of negative societal impacts include potential malicious or unintended uses (e.g., disinformation, generating fake profiles, surveillance), fairness considerations (e.g., deployment of technologies that could make decisions that unfairly impact specific groups), privacy considerations, and security considerations.
        \item The conference expects that many papers will be foundational research and not tied to particular applications, let alone deployments. However, if there is a direct path to any negative applications, the authors should point it out. For example, it is legitimate to point out that an improvement in the quality of generative models could be used to generate deepfakes for disinformation. On the other hand, it is not needed to point out that a generic algorithm for optimizing neural networks could enable people to train models that generate Deepfakes faster.
        \item The authors should consider possible harms that could arise when the technology is being used as intended and functioning correctly, harms that could arise when the technology is being used as intended but gives incorrect results, and harms following from (intentional or unintentional) misuse of the technology.
        \item If there are negative societal impacts, the authors could also discuss possible mitigation strategies (e.g., gated release of models, providing defenses in addition to attacks, mechanisms for monitoring misuse, mechanisms to monitor how a system learns from feedback over time, improving the efficiency and accessibility of ML).
    \end{itemize}
    
\item {\bf Safeguards}
    \item[] Question: Does the paper describe safeguards that have been put in place for responsible release of data or models that have a high risk for misuse (e.g., pretrained language models, image generators, or scraped datasets)?
    \item[] Answer: \answerNA{} 
    \item[] Justification: Our paper does not present any of these risks.
    \item[] Guidelines:
    \begin{itemize}
        \item The answer NA means that the paper poses no such risks.
        \item Released models that have a high risk for misuse or dual-use should be released with necessary safeguards to allow for controlled use of the model, for example by requiring that users adhere to usage guidelines or restrictions to access the model or implementing safety filters. 
        \item Datasets that have been scraped from the Internet could pose safety risks. The authors should describe how they avoided releasing unsafe images.
        \item We recognize that providing effective safeguards is challenging, and many papers do not require this, but we encourage authors to take this into account and make a best faith effort.
    \end{itemize}

\item {\bf Licenses for existing assets}
    \item[] Question: Are the creators or original owners of assets (e.g., code, data, models), used in the paper, properly credited and are the license and terms of use explicitly mentioned and properly respected?
    \item[] Answer: \answerYes{} 
    \item[] Justification: All prior work that has been used has been cited.
    \item[] Guidelines:
    \begin{itemize}
        \item The answer NA means that the paper does not use existing assets.
        \item The authors should cite the original paper that produced the code package or dataset.
        \item The authors should state which version of the asset is used and, if possible, include a URL.
        \item The name of the license (e.g., CC-BY 4.0) should be included for each asset.
        \item For scraped data from a particular source (e.g., website), the copyright and terms of service of that source should be provided.
        \item If assets are released, the license, copyright information, and terms of use in the package should be provided. For popular datasets, \url{paperswithcode.com/datasets} has curated licenses for some datasets. Their licensing guide can help determine the license of a dataset.
        \item For existing datasets that are re-packaged, both the original license and the license of the derived asset (if it has changed) should be provided.
        \item If this information is not available online, the authors are encouraged to reach out to the asset's creators.
    \end{itemize}

\item {\bf New assets}
    \item[] Question: Are new assets introduced in the paper well documented and is the documentation provided alongside the assets?
    \item[] Answer: \answerYes{} 
    \item[] Justification: We have provided details on training, as well as code and documentation.
    \item[] Guidelines:
    \begin{itemize}
        \item The answer NA means that the paper does not release new assets.
        \item Researchers should communicate the details of the dataset/code/model as part of their submissions via structured templates. This includes details about training, license, limitations, etc. 
        \item The paper should discuss whether and how consent was obtained from people whose asset is used.
        \item At submission time, remember to anonymize your assets (if applicable). You can either create an anonymized URL or include an anonymized zip file.
    \end{itemize}

\item {\bf Crowdsourcing and research with human subjects}
    \item[] Question: For crowdsourcing experiments and research with human subjects, does the paper include the full text of instructions given to participants and screenshots, if applicable, as well as details about compensation (if any)? 
    \item[] Answer: \answerNA{} 
    \item[] Justification: Our work does not involve crowdsourcing or human subjects.
    \item[] Guidelines:
    \begin{itemize}
        \item The answer NA means that the paper does not involve crowdsourcing nor research with human subjects.
        \item Including this information in the supplemental material is fine, but if the main contribution of the paper involves human subjects, then as much detail as possible should be included in the main paper. 
        \item According to the NeurIPS Code of Ethics, workers involved in data collection, curation, or other labor should be paid at least the minimum wage in the country of the data collector. 
    \end{itemize}

\item {\bf Institutional review board (IRB) approvals or equivalent for research with human subjects}
    \item[] Question: Does the paper describe potential risks incurred by study participants, whether such risks were disclosed to the subjects, and whether Institutional Review Board (IRB) approvals (or an equivalent approval/review based on the requirements of your country or institution) were obtained?
    \item[] Answer: \answerNA{} 
    \item[] Justification: This work does not involve research with human subjects.
    \item[] Guidelines:
    \begin{itemize}
        \item The answer NA means that the paper does not involve crowdsourcing nor research with human subjects.
        \item Depending on the country in which research is conducted, IRB approval (or equivalent) may be required for any human subjects research. If you obtained IRB approval, you should clearly state this in the paper. 
        \item We recognize that the procedures for this may vary significantly between institutions and locations, and we expect authors to adhere to the NeurIPS Code of Ethics and the guidelines for their institution. 
        \item For initial submissions, do not include any information that would break anonymity (if applicable), such as the institution conducting the review.
    \end{itemize}

\item {\bf Declaration of LLM usage}
    \item[] Question: Does the paper describe the usage of LLMs if it is an important, original, or non-standard component of the core methods in this research? Note that if the LLM is used only for writing, editing, or formatting purposes and does not impact the core methodology, scientific rigorousness, or originality of the research, declaration is not required.
    \item[] Answer: \answerNA{} 
    \item[] Justification: LLMs were not used in any important, original or non-standard components of this reserach.
    \item[] Guidelines:
    \begin{itemize}
        \item The answer NA means that the core method development in this research does not involve LLMs as any important, original, or non-standard components.
        \item Please refer to our LLM policy (\url{https://neurips.cc/Conferences/2025/LLM}) for what should or should not be described.
    \end{itemize}

\end{enumerate}

\end{document}